\pdfoutput=1

\documentclass[a4paper,11pt]{article}

\usepackage{lipsum}
\usepackage{tikz}
\usetikzlibrary{trees,shapes}
\usepackage{xcolor}
\usepackage{epigraph}

\definecolor{hidden-draw}{RGB}{20,68,106}
\definecolor{hidden-pink}{RGB}{255,245,247}
\definecolor{paired-light-yellow}{HTML}{FFFF88}
\definecolor{paired-light-blue}{HTML}{CCE5FF}
\definecolor{paired-light-orange}{HTML}{FFCC99}
\definecolor{paired-dark-yellow}{HTML}{FFF2CC}
\definecolor{paired-light-pink}{HTML}{FFCCCC}
\definecolor{paired-cyan}{HTML}{D5E8D4}
\definecolor{paired-gray}{HTML}{eeeeee}
\definecolor{paired-green}{HTML}{cdeb8b}
\definecolor{paired-blue}{HTML}{dae8fc}
\definecolor{paired-dark-cyan}{HTML}{a2e6eb}
\definecolor{paired-dark-pink}{HTML}{e7b2d2}
\definecolor{paired-purple}{HTML}{9999ff}
\definecolor{paired-pink}{HTML}{cc99ff}
\definecolor{paired-orange}{HTML}{ffcc99}

\definecolor{a1}{RGB}{241,233,191}
\definecolor{a2}{RGB}{255,241,218}

\definecolor{a3}{RGB}{255,239,213}
\definecolor{a4}{RGB}{250,235,215}
\definecolor{a5}{RGB}{255,239,219}
\definecolor{a6}{RGB}{255,246,225}
\definecolor{a7}{RGB}{246,227,201}
\definecolor{a8}{RGB}{254,235,226}
\definecolor{a9}{RGB}{247,220,111}
\definecolor{a10}{RGB}{199,211,189}
\definecolor{a11}{RGB}{209,196,233}
\definecolor{a12}{RGB}{214,234,248}
\definecolor{a13}{RGB}{232,245,233}
\definecolor{a14}{RGB}{237,248,177}
\definecolor{a15}{RGB}{255,228,225}
\definecolor{a16}{RGB}{255,228,181}
\definecolor{a17}{RGB}{255,222,173}
\definecolor{a18}{RGB}{255,218,185}
\definecolor{a19}{RGB}{255,203,164}
\definecolor{a20}{RGB}{247,202,201}

\definecolor{a21}{RGB}{241,254,255}
\definecolor{a22}{RGB}{230,252,252}
\definecolor{a23}{RGB}{179,236,255}
\definecolor{a24}{RGB}{174,226,249}
\definecolor{a25}{RGB}{208,234,246}
\definecolor{a26}{RGB}{189,226,219}
\definecolor{a27}{RGB}{177,204,201}

\definecolor{a28}{RGB}{216,195,216}
\definecolor{a29}{RGB}{195,155,211}
\definecolor{a30}{RGB}{208,152,223}
\definecolor{a31}{RGB}{255,183,209}
\definecolor{a32}{RGB}{255,167,209}
\definecolor{a33}{RGB}{254,235,167}
\definecolor{a34}{RGB}{255,222,137}
\definecolor{a35}{RGB}{254,180,154}
\definecolor{a36}{RGB}{247,148,161}
\definecolor{a37}{RGB}{239,154,154}
\definecolor{a38}{RGB}{255,130,171}
\definecolor{a39}{RGB}{255,105,180}
\definecolor{a40}{RGB}{251,142,172}

\usepackage{acl}
\usepackage[edges]{forest}
\usepackage{lipsum}
\usepackage{tikz}
\usetikzlibrary{trees,shapes}

\usepackage[font=small,labelfont=bf]{caption}
\usepackage{array,ragged2e}
\newcolumntype{P}[1]{>{\RaggedRight\arraybackslash}p{#1}}
\usepackage{longtable}
\usepackage{tablefootnote}

\usepackage{url}            
\usepackage{booktabs}       
\usepackage{amsfonts}       
\usepackage{nicefrac}       
\usepackage{microtype}      
\usepackage{xcolor}         
\usepackage{amsmath}
\usepackage{graphicx}
\usepackage{enumitem}
\usepackage{multirow}
\usepackage{subcaption}
\usepackage{listings}
\usepackage{tcolorbox}
\usepackage{xspace}
\usepackage{makecell}
\usepackage{soul}               
\setstcolor{red}            

\newcommand*{\Scale}[2][4]{\scalebox{#1}{$#2$}}%

\usepackage{algorithm} 
\usepackage{algpseudocode}

\usepackage{listings}
\usepackage{dirtytalk}

\usepackage{times}
\usepackage{latexsym}

\usepackage[T1]{fontenc}

\usepackage[utf8]{inputenc}

\usepackage{microtype}

\usepackage{inconsolata}

\usepackage[draft,textsize=footnotesize,textwidth=15mm]{todonotes}

\usepackage{amsfonts}
\usepackage{amsmath}

\newcommand{\queryvec}{\mathbf{q}}

\newcommand{\R}{\mathbb{R}}

\usepackage{graphicx}
\usepackage{graphics}

\title{The What, Why, and How of Context Length Extension Techniques in Large Language Models -- A Detailed Survey}

\newcommand*{\affaddr}[1]{#1}
\newcommand*{\affmark}[1][*]{\textsuperscript{#1}}
\newcommand*{\email}[1]{\texttt{#1}}
\author{Saurav Pawar\affmark[1],
S.M Towhidul Islam Tonmoy\affmark[2], S M Mehedi Zaman\affmark[2], \bf{Vinija Jain\affmark[3,4]}\footnotemark[1]\,\,, \\
 \bf{Aman Chadha\affmark[3,4]}\thanks{\,\,\,Work does not relate to position at Amazon.}\,\,, \bf{Amitava Das\affmark[5]}  \\
\affaddr{\affmark[1]Technology Innovation Institute, UAE \\
\affmark[2]Islamic University of Technology, Bangladesh \\
\affmark[3]Stanford University, USA, 
\affmark[4]Amazon GenAI, USA}\\
\affmark[5]AI Institute, University of South Carolina, USA \\
\email{saurav.pawar@tii.ae}
}

\begin{document}
\maketitle
\begin{abstract}

The advent of Large Language Models (LLMs) represents a notable breakthrough in Natural Language Processing (NLP), contributing to substantial progress in both text comprehension and generation. However, amidst these advancements, it is noteworthy that LLMs often face a limitation in terms of context length extrapolation. Understanding and extending the context length for LLMs is crucial in enhancing their performance across various NLP applications. In this survey paper, we delve into the multifaceted aspects of exploring why it is essential, and the potential transformations that superior techniques could bring to NLP applications. We study the inherent challenges associated with extending context length and present an organized overview of the existing strategies employed by researchers. Additionally, we discuss the intricacies of evaluating context extension techniques and highlight the open challenges that researchers face in this domain. Furthermore, we explore whether there is a consensus within the research community regarding evaluation standards and identify areas where further agreement is needed. This comprehensive survey aims to serve as a valuable resource for researchers, guiding them through the nuances of context length extension techniques and fostering discussions on future advancements in this evolving field.
\end{abstract}

\section{Introduction}

\textit{\say{For me context is the key - from that comes the understanding of everything}} - {\textit{Kenneth Noland}} \\

The success stories of Large Language Models (LLMs) are ubiquitous, with the advent of modern LLMs significantly advancing numerous Natural Language Processing (NLP) challenges and reaching unprecedented heights. The natural progression of scientific endeavors is to push towards new and challenging horizons. Among the ambitious initiatives, one notable effort is the extension of LLMs' understandability to encompass very long contexts. OpenAI has introduced the concept of 128 pages of context understandability, while Anthropic has recently proposed an even longer context of over 200 pages. However, a notable absence of scientific rigor is observed in these commercial releases and announcements. Several questions arise in this context: (a) What applications necessitate the understanding of such extended contexts? (b) How can we effectively measure the improved performance of applications when LLMs comprehend much longer contexts? (c) While attention mechanisms are well-studied in NLP, is there a need to devise a new form of attention specifically tailored for longer contexts?

The integration of advanced techniques designed to handle long contexts holds the potential to reshape the landscape of language models. Improved methodologies for managing long contexts could lead to increased model performance, resulting in more accurate and nuanced language understanding. Such advancements are anticipated to enhance the model’s ability to capture long-range dependencies, improving its overall effectiveness across various language tasks like:
\tikzstyle{my-box}=[
    rectangle,
    draw=hidden-draw,
    rounded corners,
    text opacity=1,
    minimum height=1.5em,
    minimum width=40 em,
    inner sep=2pt,
    align=center,
    fill opacity=.5,
    line width=0.8pt,
]
\tikzstyle{leaf}=[my-box, minimum height=1.5em,
    fill=hidden-pink!80, text=black, align=center,font=\normalsize,
    inner xsep=2pt,
    inner ysep=4pt,
    line width=0.8pt,
]
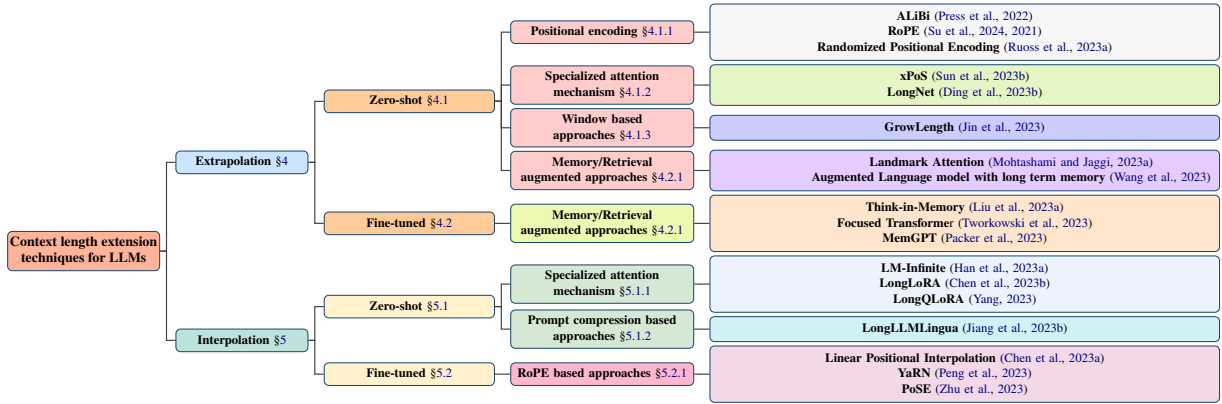
\begin{figure*}[t!]
    \centering
    \resizebox{\textwidth}{!}{
        \begin{forest}
            forked edges,
            for tree={
                grow=east,
                reversed=true,
                anchor=base west,
                parent anchor=east,
                child anchor=west,
                base=center,
                font=\large,
                rectangle,
                draw=hidden-draw,
                rounded corners,
                align=center,
                text centered,
                minimum width=5em,
                edge+={darkgray, line width=1pt},
                s sep=3pt,
                inner xsep=2pt,
                inner ysep=3pt,
                line width=0.8pt,
                ver/.style={rotate=90, child anchor=north, parent anchor=south, anchor=center},
            },
            where level=1{text width=10em,font=\normalsize,}{},
            where level=2{text width=13em,font=\normalsize,}{},
            where level=3{text width=14em,font=\normalsize,}{},
            where level=4{text width=24em,font=\normalsize,}{},
            where level=5{text width=10em,font=\normalsize,}{},
            [\textbf{Context length extension}\\ \textbf{techniques for LLMs}, for tree={fill=a35}
                [\textbf{Extrapolation} \S\ref{sec:Extrapolation}, for tree={fill=paired-light-blue}
                    [\textbf{Zero-shot} \S\ref{subsec:Zero-shot extrapolation2}, for tree={fill=paired-light-orange}
                        [\textbf{Positional encoding} \S\ref{subsubsec:Position encodings}, for tree={fill=paired-light-pink}
                        [\textbf{ALiBi} \cite{press2022train} \\ 
                        \textbf{RoPE} \cite{SU2024127063,su2021roformer} \\
                        \textbf{Randomized Positional Encoding} \cite{ruoss-etal-2023-randomized} \\
                           , leaf , for tree={fill=paired-gray}
                        ]]
                        [\textbf{Specialized attention}\\ \textbf{mechanism} \S\ref{subsubsec:Specialized attention mechanism2}, for tree={fill=paired-light-pink}
                        [\textbf{xPoS} \cite{sun-etal-2023-length} \\
                        \textbf{LongNet} \cite{ding2023longnet} \\ ,leaf, for tree={fill=paired-green}]]
                    [\textbf{Window based} \\ \textbf{approaches} \S\ref{subsubsec:SWindow based approaches}, for tree={fill=paired-light-pink}
                    [\textbf{GrowLength} \cite{jin2023growlength} \\ ,leaf, for tree={fill=paired-purple}]]
                    [\textbf{Memory/Retrieval} \\ \textbf{augmented approaches} \S\ref{subsubsec:Memory/Retrieval augmented approaches}, for tree={fill=paired-light-pink}
                    [\textbf{Landmark Attention} \cite{mohtashami2023landmark} \\
                    \textbf{Augmented Language model with long term memory} \cite{wang2023augmenting} \\ 
                    ,leaf, for tree={fill=paired-pink}
                    ]]                   
                    ]
                    [\textbf{Fine-tuned} \S\ref{subsec:Fine-tuned extrapolation2}, for tree={fill=paired-light-orange}
                        [\textbf{Memory/Retrieval}\\ \textbf{augmented approaches} \S\ref{subsubsec:Memory/Retrieval augmented approaches}, for tree={fill=a14}
                        [\textbf{Think-in-Memory} \cite{liu2023thinkinmemory} \\
                        \textbf{Focused Transforme}r \cite{tworkowski2023focused} \\
                        \textbf{MemGPT} \cite{packer2023memgpt} \\
                            , leaf, for tree={fill=paired-orange}
                        ]]]]
                [\textbf{Interpolation} \S\ref{sec:Interpolation}, for tree={fill=a26}
                    [\textbf{Zero-shot} \S\ref{subsec:Zero-shot extrapolation3}, for tree={fill=paired-dark-yellow}
                        [\textbf{Specialized attention}\\ \textbf{mechanism} \S\ref{subsubsec:Specialized attention mechanism}, for tree={fill=paired-cyan}
                        [\textbf{LM-Infinite} \cite{han2023lminfinite} \\
                            \textbf{LongLoRA} \cite{chen2023longlora} \\
                            \textbf{LongQLoRA} \cite{yang2023longqlora} \\
                            , leaf, for tree={fill=paired-blue}
                        ]]
                        [\textbf{Prompt compression based} \\ \textbf{approaches} \S\ref{subsubsec:Prompt compression based approaches}, for tree={fill=paired-cyan}
                        [\textbf{LongLLMLingua} \cite{jiang2023longllmlingua} \\ ,leaf, for tree={fill=paired-dark-cyan}]
                        ]]
                    [\textbf{Fine-tuned} \S\ref{subsec:Fine-tuned extrapolation}, for tree={fill=paired-dark-yellow}
                        [\textbf{RoPE based approaches} \S\ref{subsubsec:RoPE based approaches}, for tree={fill=a31}
                        [\textbf{Linear Positional Interpolation} \cite{chen2023extending} \\
                        \textbf{YaRN} \cite{peng2023yarn} \\
                        \textbf{PoSE} \cite{zhu2023pose} \\
                            , leaf, for tree={fill=paired-dark-pink}
                        ]]]]]
        \end{forest}}
    \caption{Taxonomy for context length extension techniques in LLMs. The figure distinguishes the techniques into interpolation and extrapolation, where they are further classified into zero-shot and fine-tuned branches. Positional encoding, Retrieval, Attention and RoPE based techniques are explored the most in this domain of context length extension.}
    \label{fig:lit_surv}
\end{figure*}
\begin{itemize}
    \item \textbf{Document Summarization:} Improved long context handling facilitates more coherent and concise document summarization, capturing essential information across extended text segments and enhancing the quality of generated summaries. A thorough understanding of the entire document, coupled with the identification of keywords and topics, necessitates adept management of an extensive contextual scope. Utilizing a shorter window in this context serves to constrict the generative capacity, potentially leading to the oversight of essential details. Furthermore, the employment of a longer contextual window proves instrumental in mitigating ambiguity, as it hinders the utilization of nuanced information without a thorough grasp of the document's intricacies. This, in turn, empowers the LLM to navigate the summarization process with heightened discernment and accuracy.
    \item \textbf{Question Answering:} The ability to consider long contexts enhances the model's comprehension of intricate question-answer relationships, resulting in more accurate and contextually relevant responses. Furthermore, LLMs exhibit enhanced proficiency in addressing QA tasks, as the resolution of co-referencing pronouns is intricately linked to the contextual entities. Additionally, when confronted with multi-turn conversations, the extension of the context window proves instrumental in facilitating the coherent tracking of the conversational topic across successive dialogues.
    \item \textbf{Language Translation:} Improved preservation of context over larger text segments enhances the model's capacity to provide accurate translations, particularly in cases where contextual nuances play a pivotal role. Polysemic lexical items present a substantial impediment in the realm of translation \cite{falkum2015polysemy}, and an augmented context window stands as a discernible aid in the contextualization of such lexemes. Furthermore, when confronted with technical jargon, LLMs exhibit enhanced efficacy when endowed with an extended input scope, particularly in accommodating domain-specific contextual nuances.
    \item \textbf{Anaphora Resolution:} Advanced handling of long contexts aids in resolving references to entities across extended text spans, contributing to more accurate anaphora resolution. The process of anaphora resolution entails establishing connections between pronouns and their respective antecedents. The extension of context windows in LLMs facilitates a more comprehensive assessment of information, thereby assisting in precise pronoun resolution through the inclusion of distant references and contextually pertinent details.
    \item \textbf{Conversational AI:} Better tracking and understanding of extended dialogues, facilitated by long context models, lead to more contextually appropriate responses in conversational AI systems. Extended context windows play a pivotal role in situating humor, sarcasm, or nuanced expressions within the conversational milieu for LLMs. This is imperative for the generation of responses that conform to the envisaged tone and stylistic nuances inherent in the ongoing dialogue.
\end{itemize}

In spite of persistent research efforts, a comprehensive overview encompassing the entire range of techniques for extrapolating context length is still absent. Additionally, the continuous evolution of LLMs has introduced innovative facets for extrapolating context length, posing challenges to existing extension methodologies and underscoring the imperative for thorough, diverse extrapolation approaches.

This paper marks the first comprehensive survey of techniques for extending the context length of LLMs. As illustrated in Figure \ref{fig:lit_surv}, we delve into existing endeavors in context length extrapolation achievable during fine-tuning. Subsequently, we delve into potential future challenges in the context length extrapolation of LLMs.

\section{Contemporary Techniques}
Several methodologies have been introduced to enhance the contextual capabilities of LLMs. For systematic categorization and enhanced clarity, we propose a taxonomy, as illustrated in Figure \ref{fig:lit_surv}. The taxonomy delineates two principal categories: Interpolation and Extrapolation techniques. Interpolation encompasses the amalgamation of information from diverse sources or contexts to refine the prediction accuracy. This technique applies to blending information originating from disparate textual segments or distinct models featuring varying context lengths. Conversely, extrapolation involves the prognostication of values beyond the confines of observed data, aiming to broaden the model's comprehension beyond its stipulated training context length. Then, there are zero-shot \cite{rashid2021towards} and fine-tuned techniques for further categorization. The rest of the subsections in the taxonomy will be discussed in the subsequent sections. 
\section{Positional Techniques}
\textls[-10]{Diverging from absolute position embeddings, relative positional embeddings are formulated based on the disparities between keys and queries \cite{shaw2018self}. A prevalent variation of relative positional embeddings was introduced in Transformer-XL \cite{dai2019transformer, yang2019xlnet}. The computation of attention scores between keys and queries has been altered to integrate trainable embeddings corresponding to relative positions. In contrast to absolute positional embeddings, Transformers equipped with relative positional embeddings showcase the capability to generalize to sequences surpassing the lengths encountered in training, demonstrating proficiency in extrapolation \cite{press2021train}. A recurring constraint associated with positional encodings is the incapacity to extend beyond the context window observed during training. Some work has been done to overcome such limitations.} \par
\textbf{Rotary Position Embedding (RoPE)} \cite{su2021roformer} employs distinct rotatory matrices based on the absolute position of each token. It calculates scores between keys and queries using relative position information, contributing to exceptional performance and prolonged decay in recent LLMs such as PaLM \cite{chowdhery2022palm} and LLaMA \cite{touvron2023llama}. \par
\textbf{Attention with Linear Biases (ALiBi)} \cite{press2021train} closely resembles T5's relative bias, introducing attention score biases penalized by distances between keys and queries. Diverging from relative positional embedding techniques like T5 \cite{raffel2020exploring}, ALiBi assigns pre-defined penalty scores without any trainable parameters. Empirical findings \cite{press2021train} indicate that ALiBi exhibits superior extrapolation performance on sequences longer than those encountered during training, surpassing various popular position embedding methods. Furthermore, ALiBi has demonstrated the capacity to enhance training stability in BLOOM \cite{scao2022bloom}. \par
\textls[-10]{\textbf{T5's relative bias} initially associates the relative separation $i-j$ of tokens located at positions $i$ and $j$ with a scalar bias value $b = f(i-j)$, where the function $f$ corresponds to a lookup table. Subsequently, the learned relative bias $b$ is incorporated into the self-attention mechanism by adding it to the dot product of the query and key. The lookup table is designed to equate distances beyond a specific threshold, ensuring adaptability to unfamiliar distances.} \par
\textls[-10]{\textbf{Position Interpolation}, introduced by \cite{chen2023extending}, is an effective method for extending context windows in pre-trained language models, particularly focusing on the LLaMA model. The key points of the method include the motivation for Position Interpolation due to sluggish adaptation in fine-tuning, the fundamental concept of scaling down position indices during pre-training, theoretical validation showcasing stability, empirical results demonstrating efficiency, an alternative approach involving attention score modification, and a fine-tuning process with robust adaptation. Comprehensive exploration shows the effectiveness of Position Interpolation in extending context windows, resulting in models proficient across diverse language tasks. Performance benchmarks indicate improved perplexity and competitive scores in passkey retrieval and long document summarization. The conclusion highlights Position Interpolation as a minimal fine-tuning method for significantly expanding context windows, providing versatile language models suitable for various applications.}\par
\textbf{Length-Extrapolatable Transformer} (LEX Transformer) has been introduced by \cite{sun-etal-2023-length}, addressing traditional Transformer limitations. It emphasizes order sensitivity, translation invariance, and length extrapolation, leveraging the Extrapolatable Position Embedding (XPOS) for a universal design with attention resolution. block-wise causal attention is introduced for improved length handling. Empirical evaluations demonstrate XPOS's consistent advantage in perplexity drop for varied lengths, with block-wise causal attention enhancing efficacy for longer sequences. The experiments underscore the crucial role of attention resolution in designing effective Transformers for diverse input lengths.\par
\textbf{Extrapolatable Position Embedding (xPos)} \cite{sun2022length} advances the Transformer's resistance to translation variations and its ability to extrapolate context length. Across each dimension of the rotational degree vector, xPos introduces a distinctive exponential decay, diminishing in magnitude as the rotation degree expands. This characteristic acts to alleviate instability during the training process, especially as the distance increases.\par
In the pursuit of extending context length, \cite{chen2023extending} and \cite{kaiokendev2023things} coincidently proposed a method that involves slight modifications to RoPE through Position Interpolation (PI) and subsequent fine-tuning on a limited dataset. As an alternative approach, \cite{ntkaware2023} suggested the "NTK-aware" interpolation, which takes into account the loss of high frequency. Subsequent developments in the "NTK-aware" interpolation method have resulted in two notable improvements, each with a specific emphasis. The "Dynamic NTK" \cite{dynamicntk2023} interpolation is designed for pre-trained models without the need for fine-tuning, while the "NTK-by-parts" \cite{bypartsntk2023} interpolation demonstrates optimal performance when fine-tuned with a small dataset featuring longer-context information.

\textbf{YaRN} \cite{peng2023yarn} diverges from Linear and NTK interpolation by implementing a ramp function, which varies the combination of Linear and NTK interpolation across different dimensions. Additionally, it incorporates a temperature factor to counteract the attention matrix's distribution shift induced by lengthy inputs.

\textbf{GrowLength} \cite{jin2023growlength} proposes a method to incrementally extend the training length throughout the pretraining phase, thereby alleviating computational expenses and improving overall efficiency. Essentially, the efficiency gains arise from training with shorter sequences and optimizing resource utilization. 
\\
\\
\textbf{Randomized Positional Encodings} \cite{ruoss2023randomized} conduct a large-scale empirical evaluation encompassing multiple algorithmic reasoning tasks, showcasing the superiority of their method compared to prior approaches. Their approach involves incorporating the positions of longer sequences by randomly selecting an ordered subset that aligns with the sequence's length.

\textbf{PoSE} \cite{zhu2023pose} propose Positional Skip-wisE (PoSE) training that smartly simulates long inputs using a fixed context window. Experimental results demonstrate that PoSE significantly reduces memory and time overhead compared to Full-length fine-tuning, with minimal impact on performance.  Exploiting this advantage, PoSE has successfully extended the LLaMA model to 128k tokens using a 2k training context window.

\textbf{LongQLoRA} \cite{yang2023longqlora} introduces LongQLoRA, an efficient and robust technique for expanding the context length of RoPE-based LLMs. Its compatibility between shift short attention and standard global attention ensures seamless integration with existing inference frameworks. With LongQLoRA, extending the context length of models like LLaMA2 7B and 13B to 8192 or 12k becomes achievable using a single V100 GPU with 32GB memory. 

\textbf{Landmark Attention}, introduced by \cite{mohtashami2023landmark}, is an innovative method to address context length limitations in Transformers by incorporating earlier input blocks directly into attention mechanisms. Using landmark tokens, the model efficiently retrieves and integrates previous blocks during inference, allowing processing of any context length. Experimental results demonstrate reduced computation cost and memory usage, showcasing the method's effectiveness for training and fine-tuning LLMs. The approach enhances interpretability, enabling a clear understanding of information retrieval. Language modeling experiments, including tasks on English books and math papers, reveal improved perplexity and the model's ability to operate effectively in longer contexts. Fine-tuning with landmark tokens extends the model's context length, outperforming the base model in passphrase recovery.

\textbf{Think-in-Memory} \cite{liu2023think} introduces TiM, a new long-term memory mechanism that mimics human memory, enabling LLMs to remember and selectively recall thoughts. TiM allows LLMs to think within the memory, eliminating the need for redundant reasoning over long-term histories.


\section{Extrapolation}
\label{sec:Extrapolation}
In this exploration, we categorize and delve into two overarching strategies: Extrapolation and Interpolation. The Extrapolation techniques aim to extend the model's comprehension to sequences beyond its initially observed lengths, employing innovative strategies to capture dependencies over extended ranges. On the other hand, Interpolation techniques concentrate on refining the model's capacity to smoothly extend its understanding of the context within the observed range, thereby enhancing performance on sequences within the initially encountered context lengths. The following sections delineate the techniques within each category, offering insights into the diverse approaches employed to address the dynamic nature of context length in LLMs.

\subsection{Zero-shot extrapolation}
\label{subsec:Zero-shot extrapolation2}
In the realm of LLMs, zero-shot context length extrapolation denotes the model's inherent capability to comprehend and generate content for input sequences of greater length than those encountered during its original training. This unique proficiency emerges without the necessity for explicit fine-tuning or additional training on lengthier sequences, showcasing the model's adaptability to extended context lengths within a given task. This capacity is of significant importance in practical applications characterized by variable input text lengths. The model, by demonstrating the aptitude to handle broader context ranges without task-specific adjustments, underscores its versatility in making meaningful predictions and generating coherent text even when confronted with contexts that surpass its training exposure. This intrinsic ability enhances the model's utility in diverse real-world scenarios where input lengths may vary, contributing to its effectiveness in processing and generating content across a spectrum of contextual complexities.

\subsubsection{Position encodings}
\label{subsubsec:Position encodings}
Position encodings emerge as pivotal components within this context, offering the model insights into the sequential structure of input sequences. By infusing information about token positions, these techniques play a foundational role in enhancing the model's ability to extrapolate its understanding to sequences of extended lengths without the need for specific fine-tuning. This section explores various position encoding techniques employed under the umbrella of zero-shot extrapolation, shedding light on their contributions to the model's adaptability to longer contexts and their impact on downstream tasks that demand nuanced comprehension of sequential dependencies.

\paragraph{Attention with Linear Biases (ALiBi)} ~

While RoPE effectively extended the context length, its limitations in zero-shot context length extrapolation were revealed by the ALiBi \cite{press2021train} research paper. 

Examining context lengths beyond those experienced in training illustrated rapid deterioration in RoPE's effectiveness. The ALiBi paper unveiled an alternate technique, highlighting superior extrapolation capabilities on their performance metrics. However, ALiBi has its drawbacks:

\begin{itemize}
    \item Its utilization of basic linear functions to modulate the attention scores across distances limits its capacity to depict intricate distance-attention functions as the Fourier basis of RoPE.
    \item Moreover, ALiBi utilizes a single function per head, reducing its capacity for expression. This may clarify why models employing ALiBi exhibit inferior performance compared to RoPE-based models on assessments such as MMLU \cite{hendrycks2020measuring} and the LMSys arena, which assesses human preferences \cite{zheng2023judging}.
\end{itemize}

\paragraph{Working of ALiBi.}


\begin{figure}
\centering
  \includegraphics[width=0.45\textwidth]{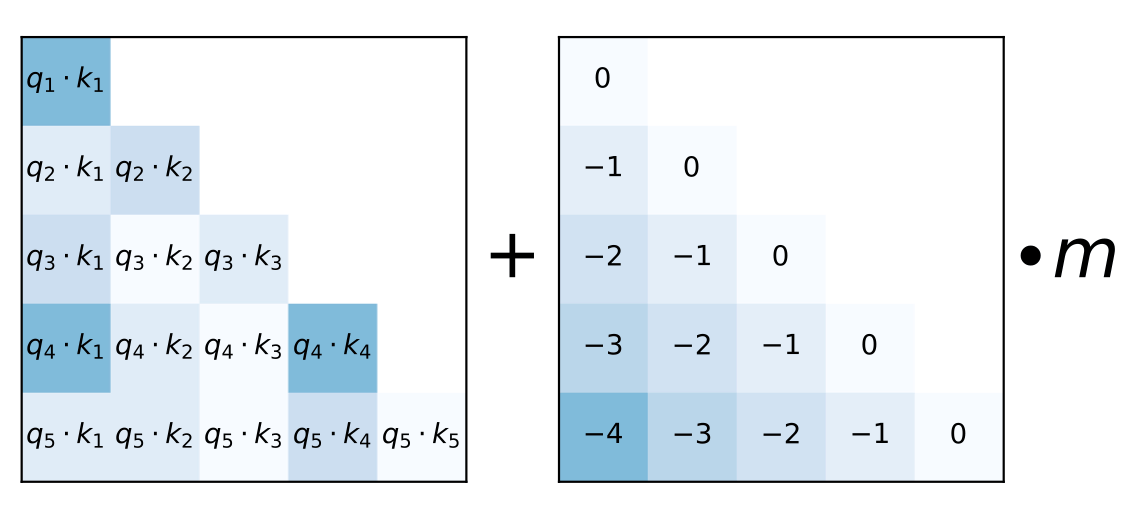}
  \caption{Implementation of ALiBi \cite{press2021train}. When calculating attention in a neural network, the figure's method involves adding a fixed bias to each attention score before applying the softmax function. This bias is the same for all attention scores in a specific head. The rest of the computation remains unchanged. The variable 'm' is a constant specific to each attention head and is set without being adjusted during training. This approach works well across different types of text, various models, and different computational resources. }
  \label{fig:AliBi}
\end{figure}
\begin{figure*}[th]
  \centering
  \includegraphics[width=0.6\linewidth]{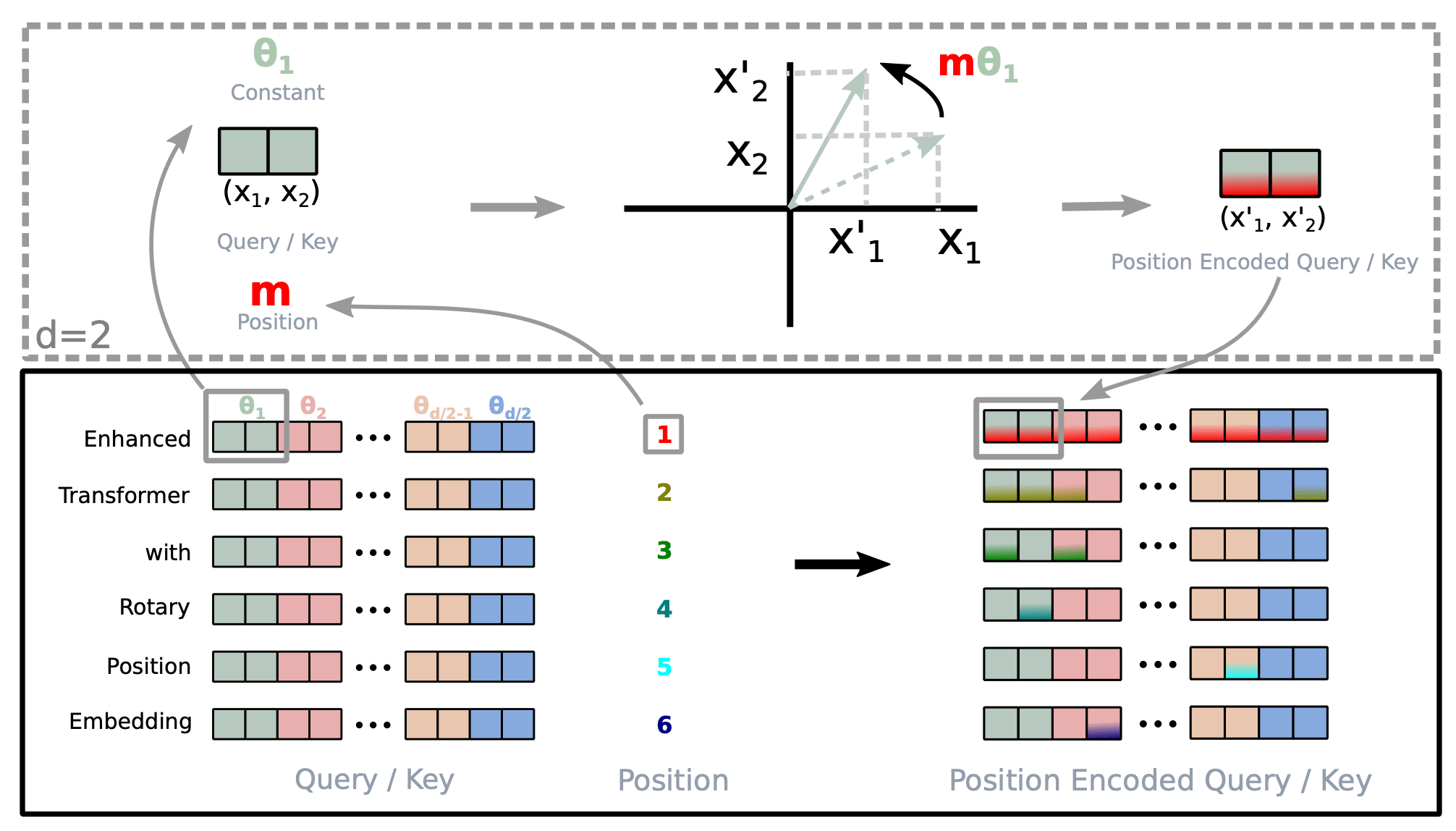}
  \caption{Visualization of RoPE \cite{SU2024127063}, which employs rotational matrices to capture precise absolute positional information in token sequences. By rotating segments of query and key projection matrices at different speeds, RoPE ensures unique rotations, influencing attention scores. The figure visually explains this innovative approach, emphasizing RoPE's reliance on relative distances for improved token relationship comprehension in self-attention models.}
  \label{fig:Rope}
\end{figure*}

For an input sequence with a length denoted as $L$, the vanilla attention layer \cite{vaswani2017attention} calculates attention scores for the $i$-th query, $\queryvec_i \in \R^{1 \times d}$ (where $1 \leq i \leq L$) in each head, based on the first $i$ keys $\mathbf{K} \in \R^{1 \times d}$, with $d$ representing the head dimension. These scores undergo multiplication by a scaling factor $\frac{1}{\sqrt{d_k}}$ and then go through a softmax function. The resulting attention scores are subsequently multiplied by the value vectors to produce the output of the attention layer. In the case of using ALiBi, no position embeddings are incorporated at any point in the network. The sole adjustment occurs after the query-key dot product, where a static, non-learned bias $m$ is added. Figure \ref{fig:Rope} offers an illustrative explanation.

\begin{equation}
\Scale[0.9]{
        \text{softmax}~(\mathbf{q}_i \mathbf{K}^{\top} + m[-(i - 1), \ldots, -2, -1, 0])}
\end{equation}

Here, $m$ is a head-specific slope, pre-determined before training. This value is essential because the dot product result between query and key can quickly escalate, so $m$ normalizes them, maintaining a range of [0,1] \footnote{The ALiBi bias is not multiplied by the scaling factor.}. \newline

\paragraph{Experiments.}

The study explores ALiBi's effectiveness on a larger model trained with a more extensive computational budget and a larger dataset (CC100+RoBERTa corpus). ALiBi exhibits robust performance comparable to the sinusoidal baseline, utilizing shorter subsequences and significantly reducing memory usage. The dataset combines RoBERTa's \cite{liu2019roberta} training corpus and the English part of the CC-100 \cite{conneau2019unsupervised} corpus (461 GB). Models with 25 transformer layers, 16 heads, and a dimension of 2048 achieve competitive perplexity with 7\% faster training and 1.6 GB less memory usage compared to the sinusoidal model. ALiBi maintains superior perplexity even when trained on sequences half the length of the baseline. Further comparisons demonstrate ALiBi's competitive performance on longer sequences, showcasing the potential for improved extended context handling. The study discusses ALiBi's efficiency in memory usage, suggesting possibilities for adding more layers.

\paragraph{Advantages.} ALiBi has garnered widespread acceptance in recent LLMs such as MPT-30B \cite{MosaicML2023Introducing}, Bloom \cite{scao2022bloom}, and BloombergGPT \cite{wu2023bloomberggpt} for the following reasons: \newline
Conventional position embeddings exhibit certain drawbacks in specific NLP applications. For instance, in cases where words exhibit nonlinear relationships with their contextual surroundings, position embeddings might fail to accurately capture these connections. Additionally, position embeddings necessitate extra computational effort, and given that they undergo learning during training, additional time may be required for optimization. In contrast, ALiBi presents a more straightforward and swifter approach that is simpler to implement and demands less computational resources. Moreover, ALiBi obviates the need for optimizing extra parameters, as the head-specific scalar bias is pre-determined and is not subjected to learning.

\paragraph{Related work.}

Concurrently to ALiBi's research, in \cite{wennberg2021case}, Wennberg et al. presented an approach involving relative positioning. Much like ALiBi's methodology, their technique introduces a bias into attention scores based on the proximity of key and query elements. In contrast, their method integrates a radial-basis function with several trainable parameters. Furthermore, their experimentation focuses on text categorization rather than language modeling, omitting any exploration of context length extrapolation.

Transformer-XL \cite{dai-etal-2019-transformer} was noted for its language model that featured a cache mechanism, expanding the inference token capacity beyond the training limits through the extension of the cache length. However, the presented results are confined to situations where the output length adheres to \textit{L} (the training length), and the method employed for relative positioning is sluggish \cite{press-etal-2021-shortformer}. In a different vein, the Longformer \cite{beltagy2020longformer} adapts models initially trained on shorter sequences for tasks at the document level. However, this adaptation entails partial training on longer sequences. The ALiBi method, on the other hand, facilitates extrapolation without the necessity for additional training on lengthier sequences.

\paragraph{Rotary Position Embedding (RoPE)} ~

The existing self-attention mechanism in pre-trained language models (PLMs), originating from the Transformer architecture \cite{vaswani2017attention}, operates without considering positional nuances \cite{yun2019transformers}. Consequently, there has been a pursuit of various methods to integrate positional information into the learning process. One method entails the inclusion of absolute position encoding derived from predetermined functions \cite{vaswani2017attention}, thereby enriching the contextual representations. Conversely, an alternative strategy involves the utilization of adaptable absolute position encoding \cite{gehring2017convolutional, devlin2018bert, lan2019albert, clark2020electra, radford2019language}. Another line of research \cite{parikh2016decomposable, huang2020improve, shaw2018self, he2020deberta, dai2019transformer, raffel2020exploring, yang2019xlnet, ke2020rethinking, huang2018music}, focuses on relative position encoding, embedding specifics about relative positions within the attention mechanism. Despite the efficacy of these approaches, they share a common trait of enhancing the context representation, which differs from the sequential self-attention arrangement. In \cite{SU2024127063}, introduce an innovative technique known as Rotary Position Embedding (RoPE) to seamlessly infuse positional information into PLMs' learning paradigm.

RoPE achieves its functionality by employing a rotational matrix to capture accurate absolute positional details, outlining the token positions relative to each other within the sequence. This process involves rotating segments of query and key projection matrices at diverse speeds, ensuring unique rotations even for tokens sharing the same encoding. Consequently, the resulting dot product varies, influencing attention scores. Discrepancies due to rotations lead to diminished dot products and attention scores, whereas alignments yield increased scores. RoPE meticulously manages these rotations for all 2-slices of query and key in the embedding dimension, establishing a nuanced attention score function across varying distances. Figure \ref{fig:Rope} offers an illustrative explanation. A key advantage of RoPE lies in its exclusive reliance on relative distances between queries and keys, eliminating the need for absolute positions. This innovative approach enhances the model's comprehension of token relationships, thereby facilitating more accurate predictions within self-attention formulations.

\paragraph{Experiments.}

The study evaluates RoFormer's performance across various NLP tasks, encompassing machine translation, pre-training with BERT, downstream assessments using GLUE benchmarks \cite{wang2018glue}, experiments involving RoPE and PerFormer's \cite{choromanski2020rethinking} linear attention. All experiments were conducted on two cloud servers equipped with $4 \times$ V100 GPUs.

In the machine translation task, the WMT 2014 English-German dataset of approximately 4.5 million sentence pairs is used. The comparison involves the transformer-based baseline from \cite{vaswani2017attention}. Modifications to \cite{vaswani2017attention} baseline enable RoPE integration during the learning process. The English-to-German translation experiment with a 37k vocabulary utilizes joint source and target byte pair encoding (BPE) by \cite{sennrich2015neural}. The evaluation, employing BLEU scores by \cite{papineni2002bleu}, consistently demonstrates RoFormer's superiority over the baseline Transformer. The PyTorch implementation with fairseq toolkit by \cite{ott2019fairseq} utilizes Adam optimizer, label smoothing (0.1), and a linearly increased and decayed learning rate. The final metric is reported from a single model averaged over the last 5 checkpoints using beam search (beam size 4, length penalty 0.6).

Pre-training experiment replaces BERT's original sinusoidal position encoding with RoPE during pre-training, utilizing the BookCorpus \cite{books2015movies} and Wikipedia Corpus \cite{wikipediacorpus2021} from the Huggingface Datasets library. The corpus is split into 8:2 train and validation sets. The evaluation metric employs the masked language-modeling (MLM) loss values, with BERT \cite{devlin2018bert} as the baseline model. In terms of implementation, RoPE is integrated into RoFormer's self-attention block. Training involves a batch size of 64, a maximum sequence length of 512 for 100k steps, and AdamW \cite{loshchilov2017decoupled} as the optimizer with a learning rate of 1e-5. Results indicate that RoFormer achieves faster convergence in MLM loss during pre-training compared to vanilla BERT.

Fine-tuning across various GLUE tasks involves the evaluation on datasets such as MRPC \cite{dolan2005automatically}, SST-2 \cite{socher2013recursive}, QNLI \cite{rajpurkar2016squad}, STS-B \cite{al2017udl}, QQP  \cite{datacanary2017quora}, and MNLI \cite{williams2017broad}, using F1-score, Spearman correlation, and accuracy as metrics. The implementation uses the Huggingface Transformers library, fine-tuning each task for 3 epochs with a sequence length of 512, batch size of 32, and learning rates 2, 3, 4, 5e-5. Results demonstrate that RoFormer significantly outperforms BERT in three out of six datasets (MRPC, STS-B, QQP).

Implementing RoPE in Performer \cite{choromanski2020rethinking}, proves effective, addressing quadratic computation costs associated with input sequence length. Tests on the Enwik8 \cite{mattmahoney2006} dataset (English Wikipedia) show improved convergence and lower loss in the 12-layer char-based Performer with 768 dimensions and 12 heads. Comparison of pre-training loss curves with and without RoPE, under consistent settings (e.g., learning rate 1e-4, batch size 128, and maximum sequence length 1024), highlights the advantages of Performer with RoPE. This implementation enhances performance while maintaining linear complexity.

\paragraph{Advantages.}
    RoPE has gained widespread adoption in recent LLMs such as PaLM \cite{chowdhery2022palm}, LLaMA \cite{touvron2023llama}, LLaMA-2 \cite{touvron2023llama2}, GPT-NeoX \cite{black2022gpt}, and Falcon \cite{almazrouei2023falcon} due to the following advantages:

    \begin{itemize}
        \item A significant advantage of rotary embeddings lies in their ability to adapt to different sequence lengths, providing flexibility in extrapolating context length. Unlike conventional position embeddings restricted to specific sequence lengths, RoPE can be adjusted to accommodate diverse sequences, making it a valuable tool for NLP models dealing with varying text lengths.
    
        \item They reduce inter-token reliance as relative distances grow, diminishing each token's impact on others as the gap widens. This is crucial for lengthy sequences, as it helps streamline computational demands while maintaining accurate predictions.
    \end{itemize}

\paragraph{Related work.}

    Lately, numerous RoPE scaling techniques have emerged to overcome RoPE's extrapolation limitations and enable its applicability to extended sequences:

    \begin{itemize}
        \item \textbf{Linear Scaling/Positional Interpolation}

        Both kaiokendev \cite{kaiokendev2023things} and Chen et al. \cite{chen2023extending} independently introduced a straightforward yet efficient method for extending context length. This technique involves dividing the position vector by a suitable scaling factor, ensuring the input fits within the original model's context length. The intuition behind this approach is to leverage the language model's interpolation capability instead of depending on extrapolation.

        \item \textbf{ReRoPE}

        ReRoPE \cite{rerope2023} expands context by altering the attention mechanism, making it more than just an embedding interpolation technique. However, it is currently not compatible with FlashAttention-2 \cite{dao2023flashattention} and necessitates two attention passes during inference.

        \item \textbf{NTK-aware RoPE scaling}

        In \cite{ntkaware2023}, the Reddit user "bloc97" introduced the "NTK-aware" interpolation method, which considers high-frequency loss. Subsequently, two enhanced versions of the "NTK-aware" interpolation have been suggested, each focusing on different aspects:

        \begin{itemize}
            \item \textbf{Dynamic NTK:}
            This \cite{dynamicntk2023} technique can be used for PLMs without the need for fine-tuning.
            \item \textbf{NTK-by-parts:}
            This \cite{bypartsntk2023} technique excels when fine-tuned with a limited quantity of long-context data.
        
        \end{itemize}

        The above NTK-aware RoPE scaling techniques have already been incorporated in open-source models such as Code-LLaMA (uses NTK-by-parts interpolation) \cite{roziere2023code} and Qwen-7B (uses Dynamic NTK interpolation) \cite{qwen}.

    \item \textbf{Truncated basis}

    In this method \cite{pal2023giraffe}, a change in the RoPE foundation involves using two cutoff values and a fixed number. The idea is to keep important elements while setting less important ones to 0. This helps the model understand longer contexts better. A fixed frequency also helps the model distinguish different distances in its training data.
    \end{itemize}

\paragraph{Randomized Positional Encodings} ~

In their work, \cite{ruoss2023randomized} illustrate that this limitation is associated with positional encodings becoming out-of-distribution for longer sequences. They introduce a novel approach to positional encodings called the randomized positional encoding scheme. This scheme mimics the positions of longer sequences by randomly selecting an ordered subset that fits the sequence's length. 

Their proposed method maintains in-domain generalization performance while significantly improving efficiency compared to the straightforward approach of training the Transformer on longer sequences. This novel family of positional encoding schemes notably enhances the length generalization capabilities of Transformers without impacting their in-domain generalization performance. 
Their large-scale empirical evaluation across various algorithmic reasoning tasks demonstrates the superiority of their method over previous approaches. The randomized encoding scheme, reliant solely on order information, exhibits remarkable performance improvements for sequences of length $M$, where $N < M \leq L$, and allows for configurable hyperparameter $L$, where $N$ represents the maximum trained sequence length. Their methodology aims to preserve the advantageous properties of relative encoding in a manner independent of the maximum training length N, enabling generalization to longer sequences during test time. Specifically, when applying their randomized positional encoding scheme, they subsample extended positions once per batch rather than individually for each sequence.

\paragraph{Working.}

When a transformer, equipped with standard positional encodings trained on a curriculum of sequences with a maximum length of N, encounters a test sequence exceeding length M > N, it leads to a redistribution of positional encodings, deviating from those observed during training. This shift becomes more pronounced as M increases.

To tackle this issue, the authors suggest a randomized encoding scheme that relies solely on order information. This scheme is anticipated to extend its applicability to sequences of length M, where $N < M \leq L$, utilizing a configurable hyperparameter $L$.

The authors assume that during each training step, the process aims to minimize the loss on a fixed-size batch of data. They define $U(S)$ as the discrete uniform distribution over set $S$, and $P_k$ as $\{S \subseteq \{1, \ldots, L\} \,|\, |S| = k\}$.

In their approach, for every training step, they begin by randomly selecting a length $n \sim U(\{1, \ldots, N\})$ (following Delétang et al., 2023) and then a random set of indices $I \sim U(P_n)$. These indices are sorted in ascending order, forming $I = \{i_1, \ldots, i_n\}$ for $i_1 < i_2 < \ldots < i_n$, ensuring no repeated sampling. The randomized positional encoding for each token $1 \leq j \leq N$ is computed as $RPE(j, \cdot) := PE(i_j, \cdot)$.

\begin{figure}
\centering
  \includegraphics[width=0.45\textwidth]{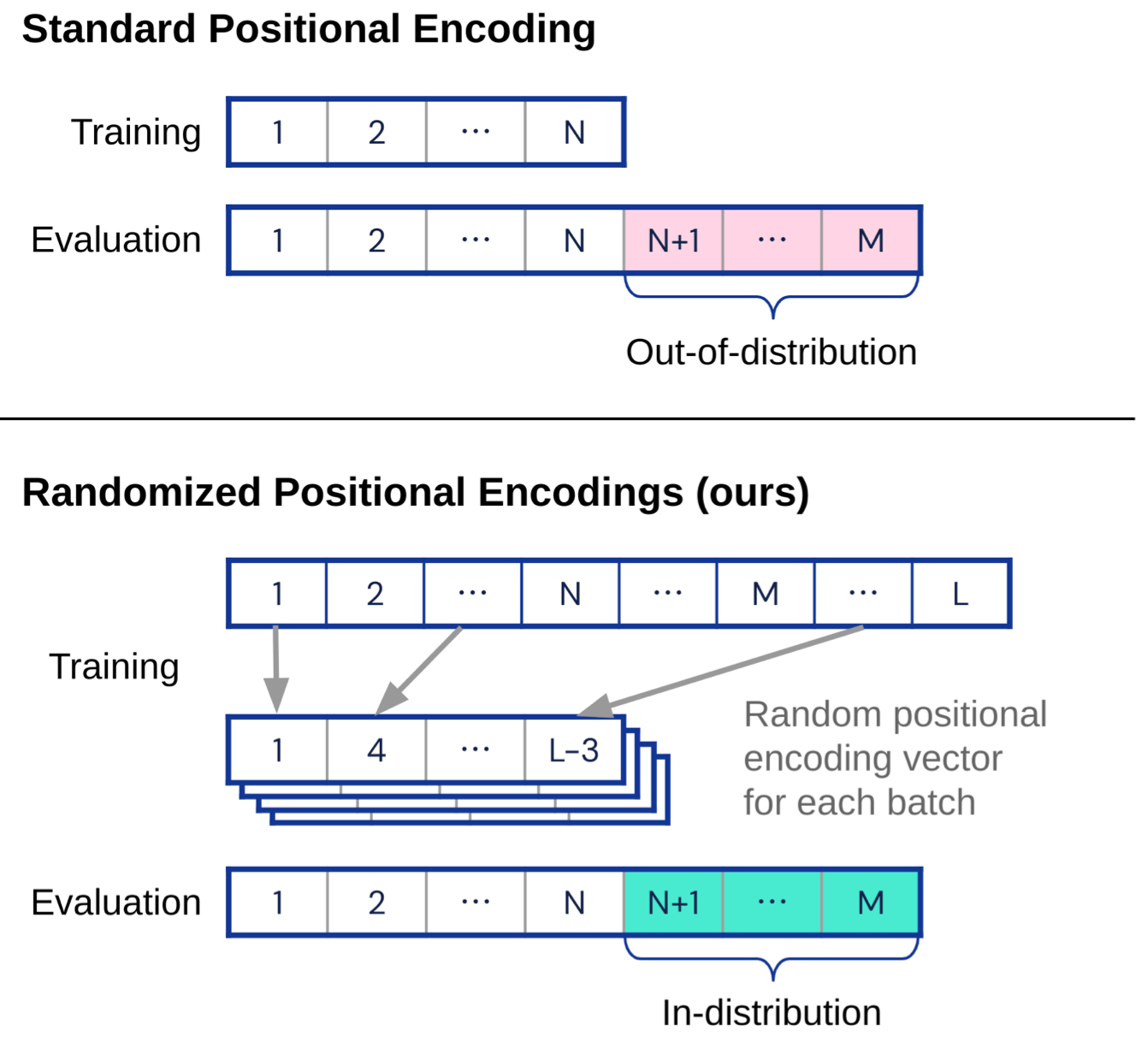}
\caption{Implementation of Randomized Positional Encodings \cite{ruoss2023randomized}. When testing a model with longer input sequences, the typical way of adding position information can lead to values that were not seen during training. The figure's solution is to address this issue by assigning a random (or ordered) positional encoding vector that covers the entire range of possible positions during testing to each training example. }
  \label{fig:RandPos}
\end{figure}

During testing, when handling sequences longer than $N$, say $M > N$, they employ a similar procedure across all token positions $1 \leq j \leq M$. This method, designed to maintain the advantageous traits of relative encoding, operates independently of the maximum training length $N$, facilitating the handling of longer sequences during testing. Figure \ref{fig:RandPos} offers an illustrative explanation.


\paragraph{Advantages.}

\textls[-10]{The proposed method exhibits superior performance in length generalization compared to previous approaches, while also demonstrating enhanced computational efficiency over the conventional method of training models on longer sequences. This method enables training on shorter sequences, achieving a test accuracy surpassing 90\%, and doing so roughly 35.4 times faster than the traditional approach of training models on longer sequences. Notably, the randomized relative encoding tackles tasks previously considered challenging by earlier methods, such as solving problems like reverse string or missing duplicate.}

\paragraph{Experiments.}

The researchers assessed the method across diverse algorithmic reasoning tasks like modular arithmetic, string manipulation (e.g., reversing/duplicating), binary operations, and bucket sorting. They evaluated their approach using \cite{deletang2022neural}'s benchmark, revealing the limitations of Transformers in generalizing to such tasks.

In their study, they utilized the encoder-only model from the original seq-to-seq Transformer \cite{vaswani2017attention}. For tasks necessitating a multi-token output sequence (e.g., string duplication), they padded the input sequence with $|y|$ empty tokens and computed the entire Transformer output based on this padded sequence.

The model underwent training on sequences uniformly sampled from $U(1, N)$, where $N = 40$, and was tested on sequences of lengths $\{N + 1, \ldots, M\}$, setting $M = 500$. The maximum position was configured to $L = 2048$. The reported accuracy averaged across all unseen sequence lengths ($N + 1, \ldots, M$), stems from the best-performing model, determined among 10 different parameter initialization seeds and utilizing three distinct learning rates: $1 \times 10^{-4}$, $3 \times 10^{-4}$, $5 \times 10^{-4}$.

\paragraph{Related work.}

Research on Transformers' positional encodings has expanded significantly. Initial methods involved simple additions of positional information, such as scaled sinusoids \cite{vaswani2017attention} or learned embeddings \cite{gehring2017convolutional}, to the input sequence embeddings. \cite{dai-etal-2019-transformer} demonstrated the benefits of incorporating relative distances between key and query vectors at each layer for enhancing the modeling of long-term inter-context dependencies.

In a similar research, \cite{su2021roformer} suggested injecting position information by rotating key-query products based on their relative distances. Additionally, \cite{press2021train} enhanced length generalization in NLP tasks by introducing a constant bias to each key-query attention score.

However, these approaches struggle with length generalization on algorithmic reasoning tasks.

\paragraph{Limitations.}

The main limitation is the need for prior knowledge of maximum test sequence length $M$ to select appropriate $L$. The evaluation is limited to synthetic algorithmic reasoning tasks, potentially lacking complexity and diversity of real applications. Additionally, a new hyperparameter, maximum sequence position $L$ is introduced. It addresses only one failure mode of Transformer length generalization on synthetic data, neglecting other factors like attention becoming less focused for extended sequences.

\subsubsection{Specialized attention mechanism}
\label{subsubsec:Specialized attention mechanism2}
Attention mechanisms stand as pivotal tools, orchestrating the nuanced allocation of importance to different segments of input sequences. These techniques empower models to dynamically focus on specific regions within the input, adapting to the varying significance of contextual information. By assigning distinct levels of attention to different parts of the sequence, these mechanisms enhance the model's ability to discern and capture relevant context, a capability crucial for tasks requiring an understanding of dependencies across diverse and extended contexts. This section explores methodologies that harness attention mechanisms, shedding light on the intricate ways in which models can selectively weigh and prioritize information within input sequences for improved context length extrapolation.

\paragraph{Length-Extrapolatable Transformer} ~

Transformers often grapple with a significant limitation: they are typically designed to handle inputs within a specific distribution size, making it impractical to train them for all conceivable input lengths. To address this, the development of a length-extrapolatable Transformer becomes imperative for broader applicability. In sequence modeling, the role of position information is pivotal in constructing accurate representations and comprehending latent meanings \cite{hochreiter1997long}. Given that various strategies focus on specific aspects of the position feature, a systematic approach is crucial for guiding Transformer design comprehensively. The proposed Transformer in \cite{sun-etal-2023-length} exhibits sensitivity to order, preventing it from devolving into a mere bag-of-words model that muddles overall meaning. Moreover, effective position translation, especially in conjunction with appropriate attention-mask operations, is essential for preserving representation integrity. Additionally, a robust sequence model must accommodate varying input lengths, a challenge unique to Transformers. The proposal introduces Extrapolatable Position Embedding (XPOS) as a universal and sound design for Transformers, leveraging the advantages of ROPE's design. Attention resolution is introduced as a metric, with the mathematical form incorporating an exponential decay in the rotation matrix to enhance position monotonicity measurement. XPOS maintains the stability of ROPE while demonstrating consistent performance in handling long-term dependencies. The incorporation of block-wise causal attention further enhances attention resolution, improving length extrapolation performance in language modeling. Training various Transformers from scratch, the LEX Transformer achieves minimal perplexity on the validation set in the pre-training corpus, validating the effectiveness of the proposed design.

\paragraph{Design architecture.}

In the exploration of Transformer model enhancements, three critical aspects are highlighted. Firstly, the model's sensitivity to order variance is emphasized to capture long-term dependencies efficiently. Position information proves crucial for effective sequence modeling, aligning with various position modeling strategies. Secondly, the concept of translation invariance is introduced, ensuring robustness in sequence representation for positional translation. This property, akin to previous work, underscores the importance of relative positions over absolute ones. Lastly, the necessity for length extrapolation capability in Transformer models is discussed. Learnable absolute position embeddings lack this ability, and various strategies show a significant drop in performance across different lengths. While approaches like Alibi address this, there is a trade-off with long-term dependency handling. The proposed XPOS is presented as a universal and robust design for Transformers, incorporating attention resolution metrics and block-wise causal attention to improve length extrapolation performance. Overall, the systematic design considerations and attention mechanisms, particularly involving relative positions, prove essential for addressing challenges in order variance, translation invariance, and length extrapolation in Transformer models.

\begin{figure}
  \includegraphics[width=0.48\textwidth]{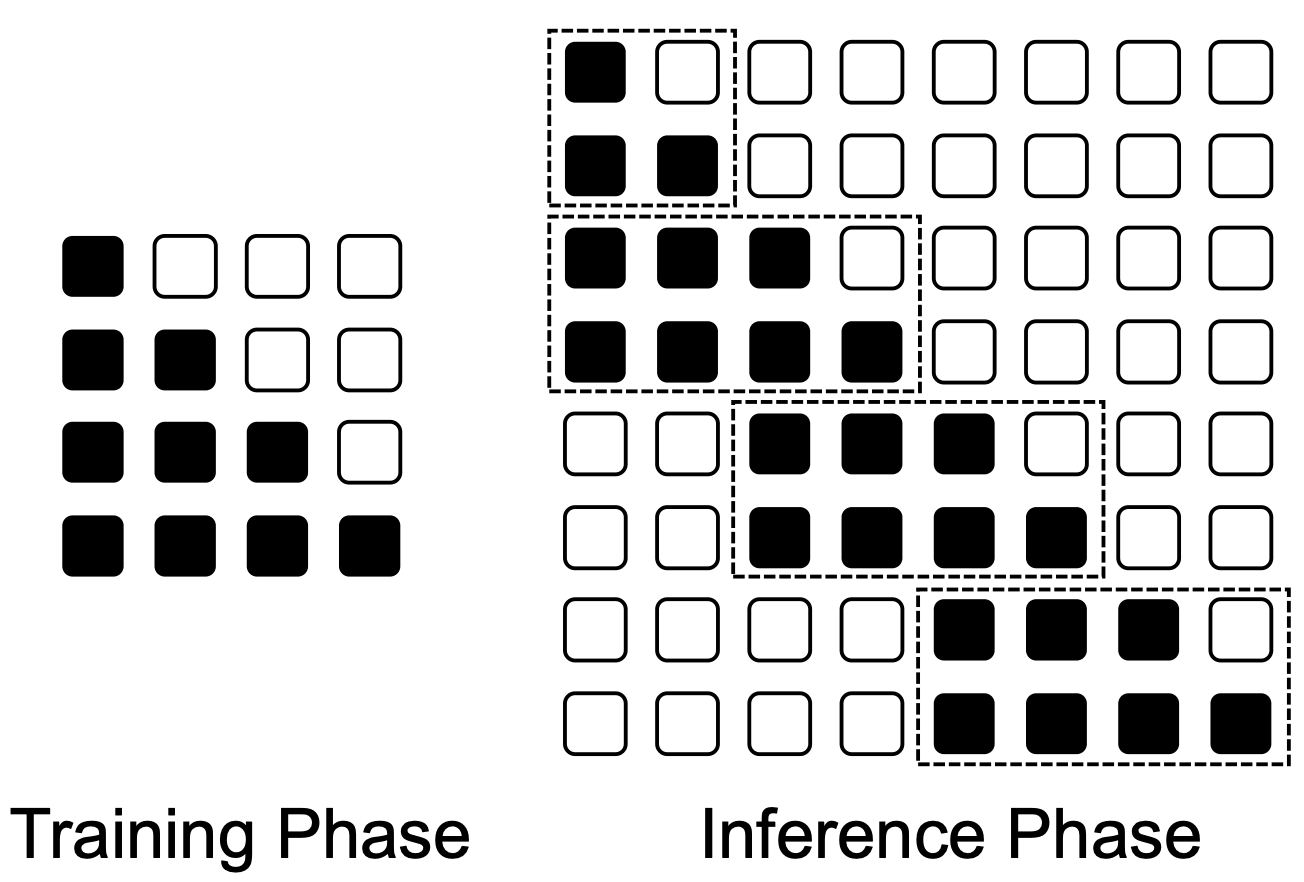}
  \caption{Implementation of block-wise Causal Attention, which is trained on short texts similar to regular Transformers, using causal masking. For longer sequences during testing, blockwise causal attention is employed, which efficiently reuses overlapping parts like key and value vectors. \cite{sun2022length}}
  \label{fig:Block-wise casual attention}
\end{figure}

To enhance the length extrapolation capabilities of Transformers, attention resolution is identified as a pivotal metric. The proposed LEX Transformer introduces two strategies to maximize attention resolution. Firstly, a method of relative position encoding, is explicitly designed for this purpose. Secondly, block-wise causal masking during inference to further improve resolution.
The essential factor for representing distance in language models is identified as the monotonicity of attention scores. The expectation of attention scores for varying distances is crucial for evaluating attention resolution.
In the pursuit of enhancing attention resolution, windowed attention strategies are explored. During inference, block-wise masking, particularly block-wise causal attention, is proposed for self-attention. This involves dividing queries into blocks during pre-training, facilitating improved resolution for encoding longer inputs. While training employs vanilla attention, the inference phase incorporates block-wise causal attention directly, contributing to enhanced position recognition, especially for longer sequences. Figure \ref{fig:Block-wise casual attention} offers an illustrative explanation. The proposed strategies collectively form the framework of the Length-Extrapolatable Transformer, offering a comprehensive approach to improving length extrapolation in Transformer models.


\paragraph{Experiments and results.}

In the experimental phase, Transformers are pre-trained from scratch with parameters resembling the medium-sized GPT-3 model, using a diverse training corpus and TorchScale framework on 16×V100 GPUs. This pre-training involves a maximal length of 1024 for memory efficiency. Language modeling evaluations, conducted on arXiv, focus on assessing the model's ability to handle long-dependency scenarios. XPOS consistently demonstrates a stable advantage in perplexity drop for lengths up to 1024, and the application of block-wise causal attention (BCA) further enhances XPOS's efficacy for lengths 2048 and 4096. The significance of resolution in effective Transformer design is confirmed through empirical evaluations measuring resolution. Specifically, the evaluation of different methods, including position embeddings (Alibi, ROPE, and XPOS) with or without block-wise causal attention, highlights the efficacy of BCA, preventing perplexity explosions in ROPE and enhancing XPOS's ability to handle longer input sequences effectively. This experiment underscores the importance of resolution in designing attention mechanisms for effective handling of long-context tasks. 

\paragraph{Related work.}
Transformers designed for extended sequences address dual challenges: inadequate efficiency in processing or memory utilization for prolonged sequences and an intrinsic balance between efficacy and resource usage. Techniques such as linear attention \cite{wang2020linformer, katharopoulos2020transformers, choromanski2020rethinking}, utilizing kernel-based or low-dimensional approximations, prioritize resource efficiency but often exhibit sub-optimal performance in typical-length scenarios. Sparse attention \cite{child2019generating, beltagy2020longformer, zaheer2020big, xiong2021simple}, leveraging structured sparsity, provides a computational reduction strategy. Additionally, designs employing recurrent-style architectures \cite{dai2019transformer, hutchins2022block, ma2022mega} for causal sequence modeling are contenders in handling these challenges. In this context, the emphasis is on addressing the extrapolation problem in language modeling—training on brief texts while evaluating extended texts \cite{press2021train}. The training methodology aligns with conventional Transformers, encompassing training on brief sequences with concentrated attention computation. The advantage lies in seamlessly unlocking the potential for long-sequence modeling during inference without compromising training efficiency. This approach guarantees the retention of optimal performance for typical lengths, eliminating trade-offs associated with long-sequence modeling compared to earlier methodologies.

\paragraph{LongNet} ~

LongNet introduces the concept of Dilated Attention, where the input $(Q, K, V)$ is segmented into sections of length $w$, denoted as $( \widetilde{Q}_i, \widetilde{K}_i, \widetilde{V}_i ) ^ \frac{N}{w}$. These segments undergo sparsification along the sequence dimension with row selection intervals of $r$. The attention computation involves parallelizing attention on the sparsified segments, which are subsequently scattered and concatenated to form the output $O$. By employing gathering and scattering operations, the dilated attention implementation seamlessly transforms into dense attention, allowing the reuse of optimizations designed for vanilla attention, such as flash attention \cite{dao2022flashattention}. This transformation results in a significant reduction in computation costs by a factor of $\frac{N}{w}r^2$ compared to vanilla attention. In practice, the segment size $w$ strikes a balance between attention's globality and efficiency, while the dilation with size $r$ minimizes computation costs by approximating the attention matrix.

\begin{figure*}[th]
  \centering
  \includegraphics[width=0.52\linewidth]{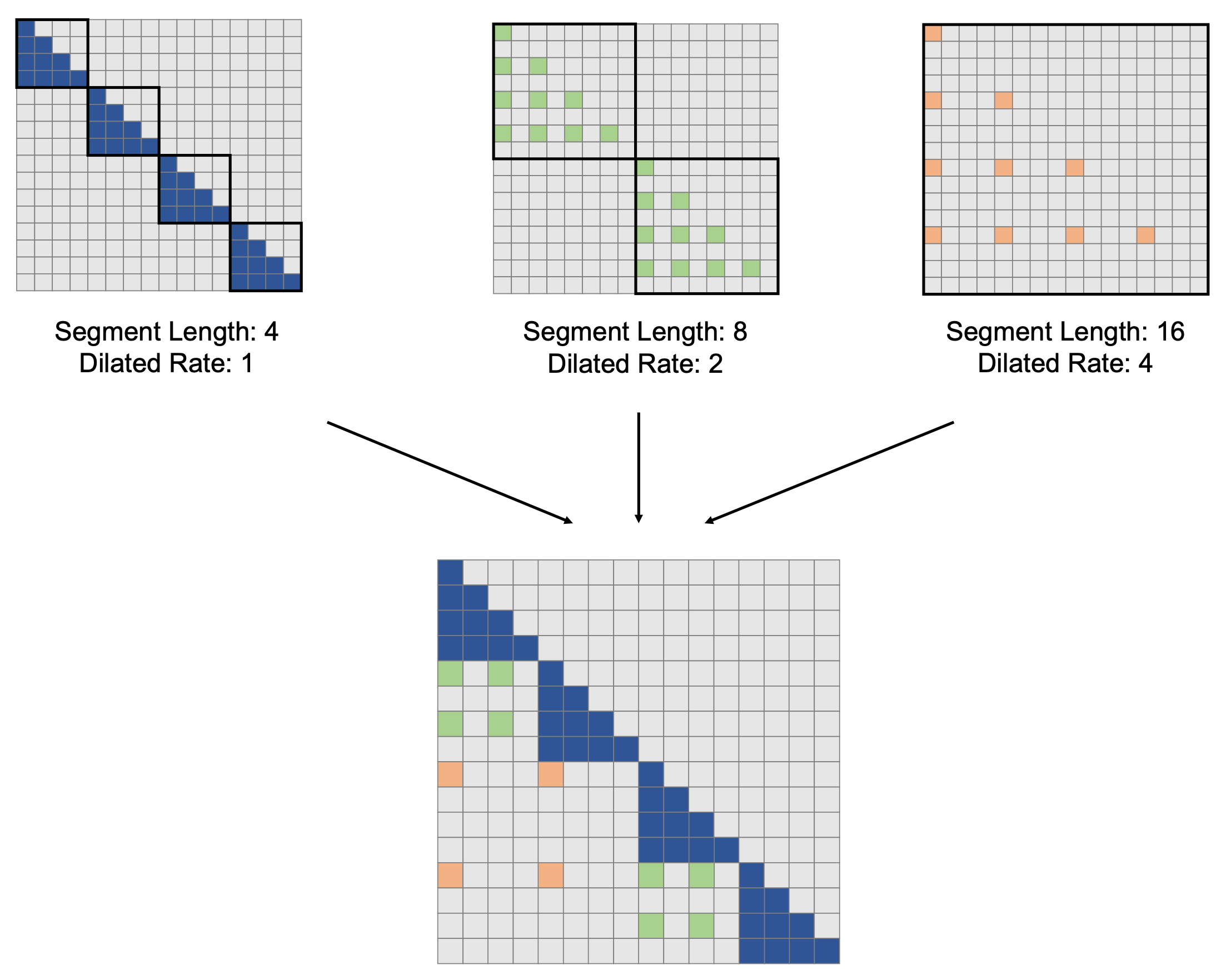}
  \caption{Implementation of Dilated Attention. The essential components of dilated attention in LONGNET \cite{ding2307longnet}, a neural network, are attention patterns designed to capture both short and long-range dependencies. The network can adjust the number of attention patterns based on the length of the input sequence. }
  \label{fig:Dilated Attention}
\end{figure*}

To capture both long-range and short-range information efficiently, LongNet utilizes a mixture of dilated attentions with various segment sizes and dilation rates $\{r_i, w_i\}^k$. Dynamic weights, calculated by the denominator of the attention softmax, are favored over fixed learnable weights, demonstrating their superior performance in experiments. The mixing of dilated attentions involves parallel computations, capitalizing on the lack of computation dependencies among them. The method progressively increases the segment size ($w_i$) and dilation rate ($r_i$) for each attention pattern until it reaches the maximum length $N$ or the predefined number of attention patterns $k$, providing an exponential attentive field. Figure \ref{fig:Dilated Attention} offers an illustrative explanation.

\paragraph{Experiments.}

In the language modeling experiments, LongNet is deployed on the MAGNETO \cite{wang2022foundation} architecture utilizing xPOS relative position encoding \cite{sun2022length}. Dilated attention replaces the standard attention mechanism while maintaining the MAGNETO base-size configuration, featuring a hidden dimension of 768, 12 attention heads, and 12 decoder layers. Pre-training is executed on The Stack dataset \cite{kocetkov2022stack}, a compilation of source code in over 300 programming languages. Data preprocessing employs the tiktoken tokenizer with $cl100k\_base$ encoding, and models undergo training with a batch size of 0.5M tokens for 300K steps. LongNet undergoes comparison with vanilla Transformer and sparse Transformers, experimenting with sequence lengths ranging from 2K to 32K. Segment lengths for LongNet are defined as $w = \{2048, 4096, 8192, 16384, 32768\}$, and dilated ratios are denoted as $r = \{1, 2, 4, 6, 12\}$. Sparse attention, following a fixed pattern \cite{child2019generating}, adjusts ratios to match computation flops with LongNet. Dense attention in vanilla Transformers is restricted to a 32K sequence length due to higher computation costs. Attention variants are derived from FlashAttention3 for training efficiency, incorporating customized flash attention kernels for both sparse and dilated attention.

For sequences surpassing the model's support, block-wise causal attention (BCA) \cite{sun2022length} is implemented for language model inference, accompanied by the removal of absolute position encoding. Results indicate that augmenting sequence length during training generally enhances language models. However, the extrapolation of sequence length in inference encounters limitations when the length significantly surpasses the model's support. LongNet consistently outperforms baseline models, affirming its efficacy in language modeling.

\paragraph{Advantages.}

LongNet presents notable advantages, encompassing linear computation complexity and a logarithmic dependency between any two tokens in a sequence. Its applicability extends to serving as a distributed trainer for exceptionally long sequences, offering a seamless integration of its dilated attention as a drop-in replacement for standard attention within the existing Transformer-based optimization framework. The linear complexity of LongNet facilitates parallelized training across nodes, overcoming computational and memory constraints through a distributed algorithm. This scalability enables efficient training on sequences of up to 1 billion tokens with nearly constant runtime, a significant improvement over the quadratic complexity limitations experienced by vanilla Transformer. This utilization of LongNet's linear computation complexity is leveraged for the distributed training of sequence dimensions.

\subsubsection{Window based approaches}
\label{subsubsec:SWindow based approaches}
In the realm of advancing LLMs, a cohort of techniques shines a light on computational efficiency by progressively extending training lengths during the pretraining phase. This strategic adaptation serves as a versatile solution to the perpetual challenge of balancing model sophistication and computational costs. By systematically increasing the training length, these techniques enhance efficiency, allowing models to grasp extended contextual nuances without imposing undue computational costs. This nuanced approach, explored in the following section, showcases a promising avenue for optimizing the practicality and effectiveness of diverse LLMs.

\paragraph{GrowLength} ~

The continuous advancement of LLMs introduces remarkable progress, albeit with heightened demands on computational resources and substantial costs. Addressing these challenges, the paper\cite{jin2023growlength} presents an innovative, straightforward, and efficient approach termed "GrowLength" to expedite the pretraining process of LLMs. The proposed method incrementally extends the training length throughout the pretraining phase, thereby alleviating computational expenses and improving overall efficiency.  This strategy empowers models to process a greater number of tokens within constrained time frames, potentially enhancing their overall performance. Essentially, the efficiency gains arise from training with shorter sequences, optimizing resource utilization. Extensive experiments with various cutting-edge LLMs demonstrate that models trained using the "GrowLength" method not only converge more rapidly but also exhibit superior performance metrics compared to those trained using existing methods. 
\begin{algorithm}[H]
\caption{Pseudocode of GrowLength}\label{alg:GrowLength}
\begin{lstlisting}[language=python]
# loader_list: data loaders with 
# different lengths.

# LLM: language model

# {nubmer}_loader: data loader for text
sequences with a length of {number}

loader_list = [128_loader,
              256_loader,...]

# Train LLMs for N epochs
for loader in loader_list:
    for batch in loader:
        loss = LLM(**batch)
        loss.backward()
        optimizer.step()
\end{lstlisting}
\end{algorithm}

\paragraph{Working.}

The core concept of GrowLength revolves around accelerated pretraining of large language models (LLMs) using shorter sequences, significantly reducing training time compared to longer sequences. Moreover, transitioning from shorter to longer sequences does not lead to a performance drop and maintains a consistent degradation trend. This method initiates pretraining with shorter sequences and gradually extends the sequence length during training, promising an efficient and seamless approach.

GrowLength extends and incorporates the context window extension technique into the pretraining stage, aiming to minimize overall pretraining time while remaining compatible with existing acceleration methods. Algorithm \ref{alg:GrowLength} offers pseudocode for the GrowLength technique. The advantages are outlined in the following section.

\paragraph{Advantages.}

\begin{itemize}
\item The method shows that training LLMs with shorter sequences is much faster than training with longer sequences.
\item When consuming the same GPU memory, training with shorter sequences allows the use of a larger batch size.
\item For smaller sequence lengths, the model can process a higher number of tokens simultaneously, exploiting the entire available memory of the GPU.
\end{itemize}

\paragraph{Related work.}

In recent research, there has been a growing interest in enhancing the efficiency of LLMs pretraining. Notable endeavors by researchers like \cite{kim2023full} have delved into optimizing CUDA kernels to minimize memory access, resulting in notable improvements in both training and inference speeds. Others, such as \cite{dao2022flashattention, choi2022accelerating, kwon2023efficient}, have explored the realms of pipeline and tensor parallelism to effectively distribute workloads across multiple GPUs, thereby enhancing the scalability of LLM inference. Additionally, strategies like quantization, investigated by \cite{wu2023understanding, dettmers2022llm, frantar2022gptq}, aim to compress LLM parameters, optimizing overall inference efficiency. These advancements play a pivotal role in addressing computational costs and time constraints in the development of new LLMs, offering valuable complements to existing methods. In the context of positional encodings within LLMs, transformer architectures have witnessed a progression from absolute positional embeddings to more contemporary approaches. Initial methods involved learnable absolute positional embeddings \cite{devlin2018bert}, providing precise position information. Subsequently, sinusoidal and fixed position embeddings were introduced to encode token positional information \cite{vaswani2017attention}. More recent innovations, like relative positional encodings, have shifted focus to leveraging distance information between tokens. \cite{press2021train} proposed a fixed linear attention bias, while \cite{su2021roformer} introduced the novel concept of rotating positive embedding (RoPE). Further advancements in extrapolation ability have been achieved through XPos \cite{sun2022length}. This diverse array of methods contributes to the evolving landscape of positional encodings in the context of LLMs.

\subsubsection{Memory/Retrieval Augmented approaches}

Memory-augmented architectures emerge as a pivotal category, introducing innovative strategies to empower models with extended contextual understanding. These approaches intricately incorporate external memory modules or mechanisms, providing the model with the ability to store and retrieve information over a broader context. By endowing models with a form of external memory, these architectures strive to enhance the retention and utilization of information beyond the immediate context window. This section explores the diverse methodologies within memory-augmented architectures, shedding light on how external memory augmentation contributes to the model's adaptability in comprehending and generating content for sequences that surpass the lengths encountered during its training phase.

\paragraph{Landmark Attention} ~

\cite{mohtashami2023landmark} introduced an innovative method to overcome context length limitations by incorporating earlier input blocks directly into attention mechanisms. The input is divided into fixed-length blocks, each marked with a landmark token acting as a gate for attention. This approach maintains random-access flexibility while providing an alternative to recurrent memory methods. During inference, attention scores on landmarks allow retrieval and integration of previous blocks, enabling processing of any context length. The method significantly reduces computation costs and memory usage. Experimental results demonstrate its effectiveness for training models from scratch or fine-tuning pre-trained models, showcasing the retrieval of information from contexts exceeding 32k tokens. The model's potential for document retrieval without additional training is also highlighted.

\begin{figure*}[th]
  \centering
  \includegraphics[width=0.8\linewidth]{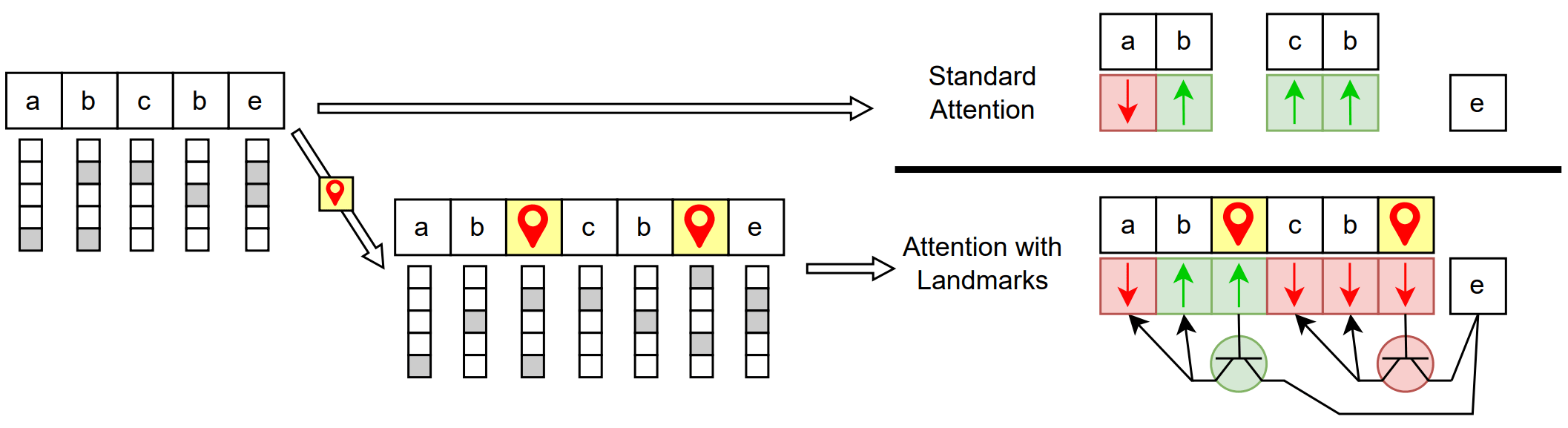}
  \caption{The comparison illustration depicts standard attention and attention with landmarks. Using a block size of $l_{\text{block}}$ = 2, it shows how a current token's attention to previous ones is influenced by similarity to both key vectors and landmark vectors corresponding to blocks. This explains why the same token can have different attention scores within different blocks, despite having the same representation initially. Landmark tokens initially share representations but evolve differently through network layers, impacting attention behavior. \cite{mohtashami2023landmark}
  }
  \label{fig:Landmark}
\end{figure*}

\paragraph{Methodology.}

This paper concentrates on causal language modeling within Transformers, where each token can only attend to preceding ones. The extension to the non-causal case is briefly discussed. While the ideal scenario for processing long inputs involves each token attending to all previous ones, this becomes computationally infeasible with increasing input length. To address this, the proposed method divides long inputs into consecutive token blocks and employs attention to retrieve relevant blocks. Representative vectors assigned to each block allow direct block retrieval based on attention scores. Landmark tokens facilitate training, and a specialized attention mechanism controls retrieval, offering semantic-based flexibility. Training details, inference processes, positional encoding, and computational advantages are discussed, showcasing the method's efficiency in handling large contexts in Transformers.

In their experiments, the researchers concentrate on assessing the efficacy of retrieving earlier blocks in language modeling tasks, specifically focusing on English language books (PG-19) and math papers from arXiv. These tasks involve long-range token interactions, making them suitable for evaluating the proposed method. The datasets used consist of 3.7 billion tokens for English books and 5.6 billion tokens for math papers from arXiv. The models trained with landmark tokens showcase the ability to retrieve relevant blocks, achieving comparable perplexity to Transformer-XL while reducing floating-point operations. Notably, the method enhances interpretability in information retrieval, allowing a clear understanding of the recovered text portions used to generate specific answers. The results also highlight that the models, employing the inference mechanism, can operate effectively in significantly longer contexts than those used during training.

The researchers employ a 12-layer GPT-2-like transformer architecture with 8 attention heads per layer, an embedding dimension of 1024, and a hidden feedforward layer size of 4096. Training uses AdamW optimizer with $\beta_1=0.9$ and $\beta_2=0.95$, weight decay of 0.001, and a cosine learning rate scheduler with warmup and minimum LR of 0.0004. GPT-2's tokenizer is used, with landmark tokens added to the dataset without altering batching. Mixed-precision training with bfloat16 is applied across up to 4 Nvidia A100 GPUs, maintaining an effective batch size of 128 via gradient accumulation and data parallelism. The model is trained for 240K steps on each dataset with context length $l_{seq}=512$. For comparison, Transformer-XL uses a window size of 256 (effective context 512) over 2048 token segments, trained for 60K steps to observe the same tokens. Figure \ref{fig:Landmark} offers a comparison between standard attention and attention with landmarks.
\paragraph{Results.}

The results assess model performance under various inference settings, particularly context lengths and block retrieval granularity. Perplexity is evaluated by dividing validation data into equally sized segments termed evaluation lengths. Segments are individually fed into the model and further divided into chunks.

\textls[-15]{Notably, a local context length of 250 tokens and retrieving the top $k=2$ most relevant blocks outperforms 512 tokens, corresponding to attending to 360 tokens (250 local context, 10 landmark, 100 retrieved). Compared to standard 360-token inference, landmark-enabled retrieval is more effective, demonstrating intelligently recovering relevant blocks allows attending to significantly fewer tokens while maintaining performance.}

Moreover, landmarks enable effective operation with longer contexts than seen during training, with perplexity improvements suggesting retrieved blocks significantly contribute, rendering results comparable to a 2048-length Transformer-XL. Unlike Transformer-XL's recurrence, the proposed method enables attending to any past token, facilitating retaining fine details and interpretability.

Performance is assessed when adjusting number of retrieved and stored blocks. With just 2 retrieved blocks at 2048 and 4096 length contexts, the model outperforms the baseline. Keeping only last 40 blocks in memory leads to better 4096-length performance, suggesting learning of Transformer-XL-like recurrence.

Additionally, cache block retrieval granularity is explored. While reducing flexibility noticeably impacts performance, the model still improves over baseline. Retrieving the same blocks (varying across heads) is possible with minimal perplexity increase.

Researchers also demonstrate fine-tuning a large language model using landmarks, extending its context length. LLaMA 7B is fine-tuned for 15,000 steps and evaluated by recovering a hidden passphrase inside text, showcasing superior generation of the correct passphrase even for much longer contexts compared to base model. When evaluating with very large inputs, additional techniques are employed to reduce memory usage by CPU offloading the KV cache except landmarks.

\paragraph{Augmenting Language Models with Long-Term Memory} ~

\cite{wang2023augmenting} introduces the Language Models Augmented with Long-Term Memory (LONGMEM) framework, designed to address memory staleness in language models. LONGMEM allows models to cache extensive previous context into a non-differentiable memory bank, employing a decoupled memory module. The approach involves a novel residual SideNet that separates the encoding of previous inputs into memory from retrieval and fusion processes, effectively mitigating memory staleness and catastrophic forgetting. By freezing the backbone LLM during efficient memory-augmented adaptation, LONGMEM taps into pre-trained knowledge without computational inefficiencies.

The LONGMEM architecture demonstrates versatility in incorporating diverse long-form text and knowledge into the memory bank based on downstream tasks. Evaluation in language modeling and memory-augmented in-context learning scenarios consistently shows LONGMEM's superiority over strong baselines. It notably improves long-context language modeling capabilities, achieving state-of-the-art performance on challenging benchmarks like ChapterBreak. Moreover, with 2k demonstration examples in memory, LONGMEM exhibits substantial in-context learning improvements on Natural Language Understanding (NLU) tasks, highlighting its efficacy in enhancing language models across various contexts and learning scenarios.

\paragraph{Design architecture.}

In the methodology section, the authors introduce the LONGMEM framework to enhance the ability of LLMs to harvest relevant information from past long contexts. They propose augmenting the frozen backbone LLM with a decoupled memory module, employing a lightweight residual SideNet for efficient training. The architecture involves three key components: the frozen backbone LLM, SideNet, and Cache Memory Bank. Previous and current inputs are encoded differently using the frozen backbone LLM, and the SideNet module acts as an efficient adaptation model, fusing current input context and caching previous contexts in a decoupled memory.

Within the Residual SideNet section, the authors detail the SideNet architecture and initialization process, emphasizing its efficiency through pre-trained parameters. Cross-network residual connections are introduced to fuse representations from the backbone LLM into SideNet, ensuring knowledge transfer from pre-trained parameters. In the Memory Retrieval and Fusion subsection, the authors describe LONGMEM's long-term memory capability achieved through a memory-augmentation module for retrieval and fusion. Token-to-chunk memory retrieval, focusing on n-gram structures, and memory fusion within a special memory-augmented layer are highlighted. The methodology outlines an efficient and innovative approach to address memory staleness and enhance the capabilities of LLMs across various downstream tasks.

\begin{figure*}[th]
  \centering
  \includegraphics[width=0.8\linewidth]{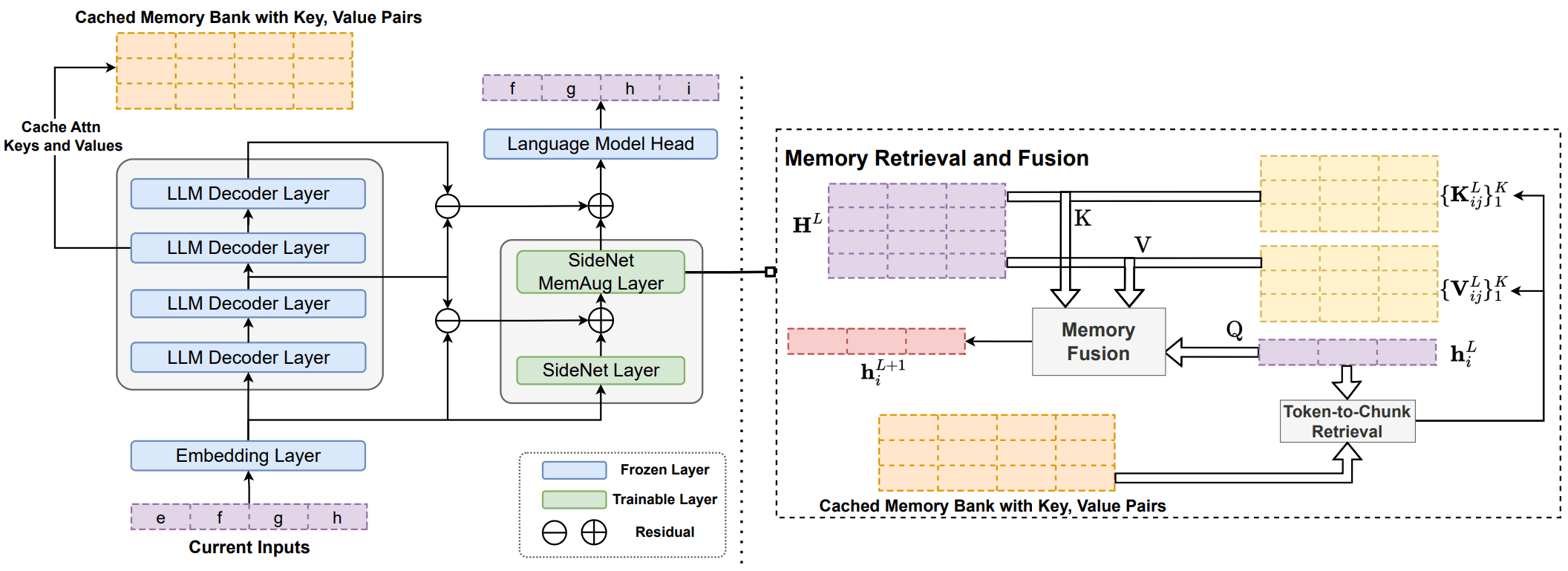}
  \caption{Overview of LongMem \cite{wang2023augmenting} architecture, where language models are enhanced to effectively use information from a long past context by adding a separate memory module to the existing model. A lightweight SideNet is also introduced to integrate memory context information efficiently. The language modeling problem and SideNet are showed, and the processes of encoding, storing, recalling, and integrating past memory for better language modeling are also portrayed in the figure.}
  \label{fig:LongMem}
\end{figure*}

\paragraph{Experiments.}

In the experimentation phase, the authors assess the performance of the proposed LONGMEM model across various tasks requiring in-memory long-contexts. The evaluation encompasses long-text language modeling and language understanding by loading past long contexts into cached memory, as well as infinite-length in-context learning achieved by loading a large number of demonstration examples into cached memory.

\textls[-10]{The training setup details the batchifying process for training corpora, emphasizing the need for maintaining global causality at the segment level. The authors sample a subset of the Pile as the training corpus, reproduce the GPT-2 (407M-params) as the pre-trained backbone LLM, and introduce the SideNet and Cache Memory Bank components. The training iterates on 26B tokens with specific hyperparameters for memory-augmented adaptation. Memory retrieval details involve constructing and updating memory retrieval modules for efficiency, utilizing token-to-chunk retrieval, and introducing baselines like GPT-2 and Memorizing Transformer (MemTRM). The subsequent subsection focuses on long-context language modeling, highlighting the benefits of augmented decoupled memory in providing significant background and contextual information. Evaluation settings for Project Gutenberg 2020-2022, arXiv, and ChapterBreak datasets are outlined, with the chosen metrics being perplexity and suffix identification accuracy. The methodology and evaluation design provides a comprehensive understanding of LONGMEM's capabilities and its comparison against relevant baselines in handling diverse tasks requiring extensive context utilization.}

\paragraph{Results.}

The LONGMEM model, as proposed, exhibits notable superiority over all considered baselines in the realm of long-text language modeling. Demonstrating improvements ranging from -1.38 to -1.62 perplexity on various length splits of the PG-22 dataset and -1.0 perplexity on arXiv datasets, the method showcases its efficacy in comprehending past long-contexts stored in cached memory for enhanced language modeling. Moreover, LONGMEM achieves state-of-the-art performance with 40.5\% accuracy on the ChapterBreakAO3 suffix identification benchmark, surpassing both strong long-context transformers and the latest LLM GPT-3, which boasts 313 times larger parameters. These substantial enhancements underscore LONGMEM's ability to adeptly utilize cached memory to complete language modeling tasks with a keen understanding of future inputs.

Moving to the domain of memory-augmented in-context learning, LONGMEM extends the capabilities of LLMs in this regard. Traditional in-context learning is constrained by input context length, limiting its effectiveness in absorbing supervision from sufficient demonstration examples. LONGMEM addresses this limitation by introducing unlimited-length memory augmentation, enabling it to attend to the entire training set by loading it into cached memory. This innovative approach goes beyond conventional few-shot in-context learning, realizing memory-augmented in-context learning with thousands of auxiliary demonstration examples. Evaluation results on various Natural Language Understanding (NLU) datasets, such as SST-2, MPQA, MR, Subj, and SST-5, demonstrate remarkable improvements in both 20-shot and 4-shot scenarios. LONGMEM exhibits an average score increase of +8.0 over pre-trained GPT-2* and MemTRM in the 20-shot setting, emphasizing its proficiency in utilizing auxiliary contextual demonstrations for superior in-context learning. Additionally, the model shows promise in open-ended generation tasks, achieving a +4.5 EM score increase on SQuAD, showcasing its versatility in leveraging cached memory for improved in-context learning. The results affirm LONGMEM's effectiveness and superiority in long-context modeling, understanding, and many-shot in-context learning, establishing it as a potent approach in the landscape of language models. Ablation studies further explore the impact of hyperparameters, such as chunk size and memory size, providing insights into their effects on task performance.

\paragraph{Related work.}

Prominent language models like GPT-2 \cite{radford2019language}, GPT-3 \cite{brown2020language}, OPT \cite{zhang2022opt}, and BLOOM \cite{workshop2022bloom} have drastically reshaped NLP research, elevating performance benchmarks in language understanding, generation \cite{wang2022task}, and vision-language \cite{wang2022visually} tasks. These models, collectively known as LLMs, exhibit groundbreaking abilities such as few-shot in-context learning and multi-step reasoning \cite{wei2022chain} by scaling their parameters. To address the challenge of processing longer contexts, a category of transformer models, termed \say{x-formers}, has been proposed. Transformer-XL \cite{dai2019transformer} pioneers a caching mechanism for attention keys and values from past segments, while recent innovations like LinFormer \cite{wang2020linformer}, LongFormer \cite{beltagy2020longformer}, and Routing Transformer \cite{roy2021efficient} leverage sparse attention mechanisms to mitigate the quadratic complexity issue. Despite their efficiency gains, these models face limitations when dealing with book-length sequences. BigBird \cite{zaheer2020big} extends sequence length but remains constrained to 16k tokens. In the realm of task-specific tuning, the Side-Tuning method \cite{zhang2020side} involves training a lightweight side network, integrated with the pre-trained network through summation. In contrast, LONGMEM introduces decoupled memory to enhance long-term input memorization without task-specific tuning. Its distinctive cross-network residual connections set it apart from the conventional summation approach in Side Tuning.

\subsection{Fine-tuned extrapolation}
\label{subsec:Fine-tuned extrapolation2}
Fine-tuned extrapolation in the context of LLMs represents a sophisticated evolution in the domain of NLP. This process involves specifically refining a model's existing capabilities to not only comprehend but also accurately generate text that extends beyond the parameters of its initial training data. Unlike zero-shot learning, where the model leverages its pre-trained knowledge without further adjustments, fine-tuned extrapolation focuses on enhancing the model's proficiency with additional, targeted training. This is particularly crucial for applications that demand high precision in generating contextually rich and nuanced text. By undergoing fine-tuning, the LLM becomes adept at handling complex and lengthy inputs, demonstrating a remarkable flexibility in adapting to new content types and structures. This heightened capability ensures that the model can produce more coherent, contextually appropriate, and sophisticated responses, thereby significantly elevating its applicability across a myriad of scenarios, from advanced conversational interfaces to comprehensive content creation. The advent of fine-tuned extrapolation marks a pivotal stride in the journey towards more intelligent, responsive, and versatile language models, capable of navigating the intricacies of human language with unprecedented finesse.

\subsubsection{Memory/Retrieval augmented approaches}
\label{subsubsec:Memory/Retrieval augmented approaches}
Two notable methods, TiM (Think-in-Memory) and Focused Transformer (FOT), have emerged to address the challenge of extending the effective context length in LLMs. TiM introduces a dynamic memory mechanism, facilitating improved performance in long-term interactions by eliminating repeated reasoning and enhancing the organization of historical thoughts. On the other hand, FOT employs a contrastive learning-inspired training process, effectively extending the (key, value) space in an attention layer with access to external memory. FOT demonstrates its efficacy through fine-tuning large-scale models, showcasing enhanced performance in tasks requiring a longer context. Both methods contribute significantly to overcoming limitations related to effective context length, offering versatile solutions for optimizing LLMs in real-world applications.

\paragraph{Think-in Memory} ~

\cite{liu2023think} introduces TiM, a new long-term memory mechanism that mimics human memory, enabling LLMs to remember and selectively recall thoughts. TiM allows LLMs to think within the memory, eliminating the need for redundant reasoning over long-term histories.

\paragraph{Working.}

The proposed TiM enables the agent to engage in long-term conversations, retaining valuable historical information across multiple interactions.

TiM comprises interconnected components aimed at enhancing coherence and accuracy in extended conversations: The first component, the \textbf{Agent}, is a pre-trained LLM model tailored for dynamic conversations. The second component is the \textbf{Memory Cache}, an expanding hash table storing key-value pairs representing individual thoughts. Lastly, the \textbf{Hash-based Mapping} incorporates locality-sensitive hashing for efficient storage and retrieval of relevant thoughts.

The TiM framework operates in two stages:
Stage-1: \textbf{Recall and Generation}: When a user asks a question, the LLM agent retrieves relevant thoughts from memory, facilitating accurate responses without redundant reasoning over raw conversation text.
Stage-2: \textbf{Post-think and Update}: Following the response, the LLM agent conducts post-thinking on the Q-R pair and integrates newly generated reasoning thoughts into the memory cache.

\paragraph{Advantages.}

The authors extensively experimented with multi-turn dialogue datasets, revealing significant enhancements in various dimensions of LLM performance:

\begin{itemize}
    \item It accommodates diverse topics, spanning from open to specific domains.

    \item It supports bilingual languages, encompassing both Chinese and English.

    \item It notably enhances response correctness and coherence.
\end{itemize}

\paragraph{Experiments.}

Three distinct datasets, including \textbf{KdConv} \cite{yang2023baichuan}, \textbf{Generated Virtual Dataset (GVD)} \cite{zhong2023memorybank}, and the manually curated \textbf{Real-world Medical Dataset (RMD)}, serve as demonstrations for the proposed method's effectiveness.

To highlight the efficacy of the TiM mechanism, the authors integrate two robust LLMs \cite{zeng2022glm, yang2023baichuan}. They employ three metrics—retrieval accuracy, response correctness, and contextual coherence—to evaluate the method. To ensure fairness during evaluation, the prediction outcomes of all LLMs are randomized before human evaluation.

Evaluations performed on both English and Chinese test sets from the GVD dataset \cite{zhong2023memorybank} show that the method outperforms SiliconFriend \cite{zhong2023memorybank} across all metrics, particularly excelling in contextual coherence, indicating the TiM mechanism's efficacy across languages.

When tested on various topics (film, music, and travel) within the KdConv dataset, the method showcases superior performance across all topics. Notably, it achieves high retrieval accuracy, mitigating lower response correctness observed in LLMs without a memory mechanism, and significantly enhances contextual coherence in responses.

On the RMD dataset, the method notably improves response correctness and contextual coherence for ChatGLM and Baichuan2 in long-term medical conversations. The approach aligns more closely with human memory workflows, enabling LLMs to generate more human-like responses.

The authors introduce a medical agent, TiM-LLM, tailored for patient-doctor conversations, combining ChatGLM and TiM. TiM-LLM assists clinical doctors by accurately recalling symptoms and comprehensively understanding patient diseases to provide precise diagnoses and treatment options.

\paragraph{Related work.}

Numerous strategies have been explored to bolster the memory capacities of LLMs. Memory-augmented networks (MANNs), such as Neural Turing Machines (NTMs) \cite{graves2014neural} and other variants like the one presented by Meng et al. \cite{meng2018dialogue}, leverage external memory caches for handling extensive context information in dialogues. These MANNs manipulate and store data, facilitating tasks requiring long-term context through memory interactions.

Several recent studies have specifically delved into long-term conversations \cite{xu2021beyond, xu2022long, zhong2023memorybank, liang2023unleashing}. For instance, \cite{xu2021beyond} introduced a new English dataset, compiling multi-session human-human crowd-worker chats to address the nuances of long-term conversational flows. In a parallel effort, \cite{zhong2023memorybank} proposed the MemoryBank mechanism, drawing inspiration from Ebbinghaus' forgetting curve theory. However, these approaches encounter significant challenges in establishing a robust and adaptable long-term memory framework for LLMs. They primarily focus on storing raw dialogue text, necessitating repeated reasoning by the LLM agent over the same historical data. Moreover, these methods involve computationally intensive pairwise similarity calculations to recall relevant information, proving time-consuming in prolonged interactions.

\paragraph{Focused Transformer} ~

The study by \cite{tworkowski2023focused} identifies a primary challenge in context augmentation: as the number of documents increases, the relevant-to-irrelevant token ratio diminishes, leading to overlaps between keys associated with irrelevant and pertinent values. This challenge, termed the distraction issue, hinders the model's ability to differentiate between them. To address this, the researchers propose the Focused Transformer (FOT), a technique designed explicitly to combat the distraction issue. FOT enables attention layers to access an external memory of (key, value) pairs through the k-nearest neighbors (kNN) algorithm, extending the total context length effectively. The training procedure, inspired by contrastive learning, exposes memory attention layers to both relevant and irrelevant keys during training, enhancing their ability to differentiate semantically diverse values. The researchers introduce fine-tuned OpenLLaMA models with FOT, demonstrating its applicability to existing models and significant improvements on tasks requiring long-context modeling.

FOT is presented as a plug-and-play extension of transformer models, applicable for both training new models and fine-tuning existing large models with longer context. Leveraging memory attention layers and a crossbatch training procedure, FOT enables the model to retrieve information from external memory during inference, extending the context effectively. The crossbatch training procedure guides the model to learn (key, value) representations conducive to memory attention layer usage. The memory attention layers access an external memory database during inference, ranking, and retrieving keys using the kNN search algorithm. The training procedure introduces a novel approach to improve the structure of the (key, value) space, inspired by contrastive learning, exposing attention layers to relevant and irrelevant keys in a differentiable manner.

\begin{figure*}[th]
  \centering
  \includegraphics[width=0.8\linewidth]{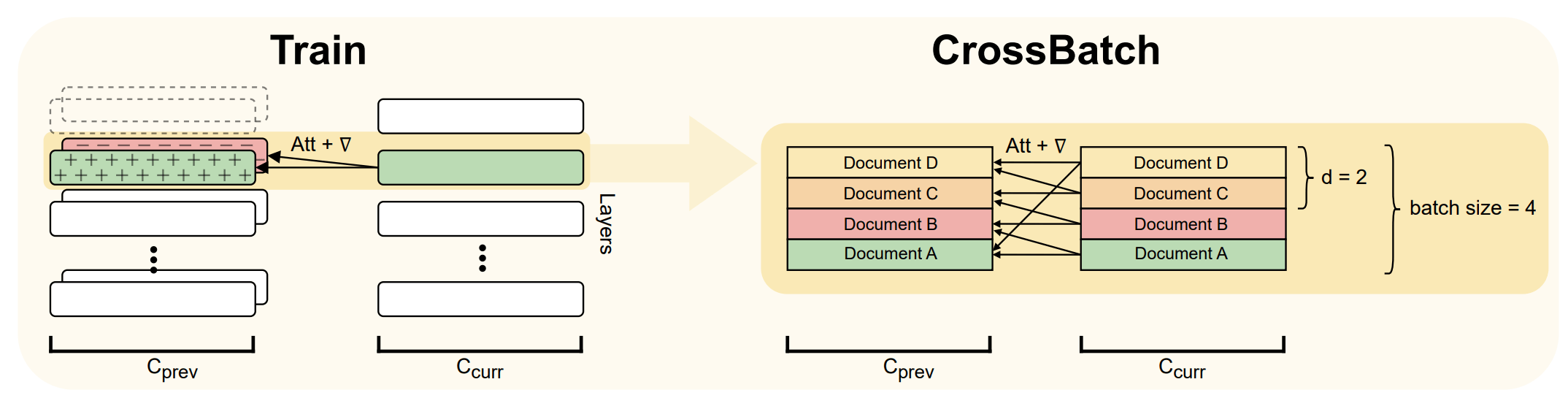}
  \caption{Overview of Focused Transformer \cite{tworkowski2023focused} during training. FOT incorporates memory attention layers and employs a crossbatch training approach. The memory attention layers allow the model to access information from additional context during inference, effectively expanding the context. }
  \label{fig:Focused transformer training}
\end{figure*}

\begin{figure}
\centering
  \includegraphics[width=0.32\textwidth]{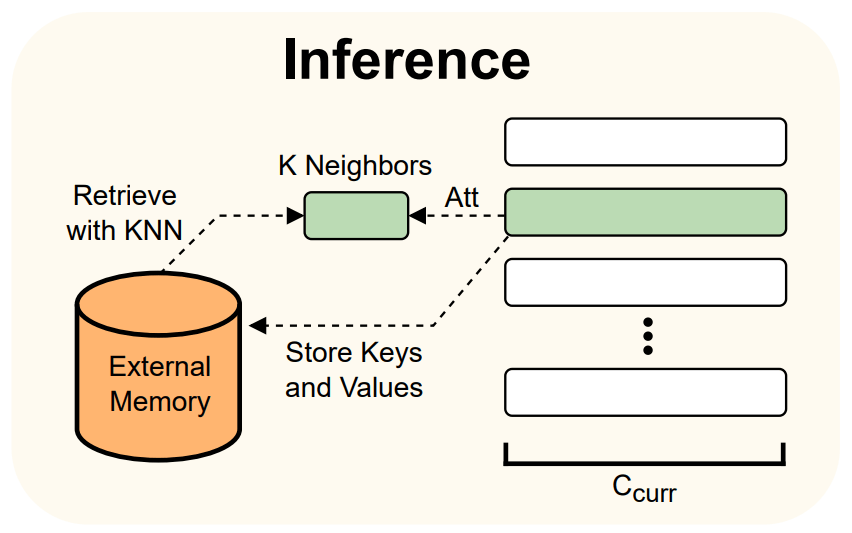}
  \caption{Overview of Focused Transformer \cite{tworkowski2023focused} during inference. During inference, the memory attention layers in FOT facilitate the retrieval of information from an extended context, enhancing the model's understanding. This is made possible by the (key, value) representations learned by the model during the training phase, guided by the crossbatch training procedure. This procedure encourages the model to acquire representations that are particularly compatible with the memory attention layer, optimizing its performance in utilizing longer context information. }
  \label{fig:Focused transformer inference}
\end{figure}

The distraction issue is addressed in the study, highlighting that during standard training, the model is not incentivized to distinguish keys from different documents, resulting in an evenly spread attention mass on related and unrelated documents. The distraction issue is mitigated through the crossbatch training procedure, leading to focused attention and improved performance. The proposed methodology is versatile, allowing for the use of external memory without requiring it during training. The study introduces minimal additional hyperparameters and demonstrates the effectiveness of FOT, particularly evident in LONGLLAMAs' significant improvements on tasks requiring long-context modeling, such as the 256k context length passkey retrieval task. Figure \ref{fig:Focused transformer training} and Figure \ref{fig:Focused transformer inference} offer an illustrative explanation of the Focused Transformer during training and inference respectively. 

\paragraph{Experiments.}

The researchers demonstrate the applicability of the FOT to fine-tune existing large models, specifically OpenLLaMA-3B and OpenLLaMA-7B models. The resulting models, termed LONGLLAMAs, exhibit the capability to extrapolate beyond their training context length, reaching up to 256K, while maintaining performance on short-context tasks. The experimental setup involves using L = {6, 12, 18} (for 3B) and L = {8, 16, 24} (for 7B) as memory layers, fine-tuning on 10B (for 3B) and 3B (for 7B) tokens with FOT, 8k context length, and a dataset mixture based on RedPajama. Noteworthy modifications include retaining positional encodings, using dense attention instead of kNN retrieval, and adjusting the crossbatch training procedure for more control.

The effective context length of LONGLLAMAs is evaluated through the passkey retrieval task, showcasing the model's ability to solve tasks beyond its training context length. Subsequent assessments focus on measuring long-context capabilities on downstream tasks, specifically TREC question classification and WebQS question answering, demonstrating significant accuracy gains with longer contexts. A comparison with standard long-context fine-tuning reveals FOT's superior performance in accuracy improvements, particularly when evaluated beyond the training length.

Importantly, the researchers emphasize that fine-tuning for longer contexts with FOT does not compromise performance on short-context tasks, ensuring compatibility and supporting the use of LONGLLAMAs as drop-in replacements for original LLaMA models. The research provides valuable insights into the effectiveness and versatility of FOT in extending context lengths and enhancing model performance across various tasks.

\paragraph{Analysis.}

In this section, the researchers conduct comprehensive experiments on smaller models to scrutinize and validate their approach further. The investigation addresses key questions: (1) The performance of FOT when scaling context length at inference, (2) FOT's capability to extend the context length of existing pre-trained models, and (3) its effectiveness in handling distractions and its impact on performance in long-context language modeling tasks. Additionally, ablation studies and further analyses are presented.

The experimental setup involves decoder-only Transformer models with 12 layers and 184M parameters, using $l$ = 8 as the memory attention layer and tuning k = 128 for the top keys retrieved by kNN. Two evaluation settings, single-document, and multi-document, are distinguished. The datasets evaluated include PG-19 (English books), arXiv (mathematical papers), GitHub (code), and Isabelle (formal proofs).

In the analysis of scaling context length to 16M, a synthetic dictionary lookup task is employed. FOT is compared to a baseline transformer model, demonstrating FOT's effectiveness in utilizing large memory for extended context length.

FOT's fine-tuning capability and context length extrapolation are explored. Perplexity improvements on various datasets are studied, showcasing FOT's ability to enhance performance even beyond the training context length. A comparison with baselines reveals steady perplexity gains, emphasizing FOT's advantages.

The section delves into distractions in language modeling tasks, particularly in the multi-document setting. Using the PG-19 dataset, the researchers measure perplexity variations with different multi-doc memory sizes. The findings indicate that higher values of the crossbatch dimension result in improved perplexity, aligning with earlier observations on mitigating the distraction issue.

context length extrapolation in the single-doc setting is explored, revealing that FOT aids in extrapolating to longer contexts, even beyond the training context length. The analysis introduces an additional parameter w, showing improvements as context grows.

Ablation studies focus on two key properties: differentiability and the inclusion of negatives. Differentiable keys and values are compared to the Memorizing Transformer, affirming the benefits of FOT. The importance of negatives is underscored, showing their significance in achieving better model performance.

The section concludes by discussing the relation to Memorizing Transformer, emphasizing the impact of training protocols and memory integration approaches. FOT's simplicity in memory integration is highlighted, along with a proof-of-concept experiment combining training protocols. The study recommends FOT's approach due to its ease of fine-tuning existing models and potential benefits in training protocols.

\paragraph{Limitations and posssile remedies.}

The current study not only provides insights into the challenges and advancements in the development of the Focused Transformer but also identifies several areas for future exploration and potential improvements. The following outlines the avenues for future research and acknowledges existing limitations:

Memory Scale: A crucial direction for future research involves scaling up memory capacity. Overcoming engineering challenges to store more than 16 million (key, value) pairs will necessitate the implementation of distributed multi-node systems. While the experiments utilized exact kNN search, which is limited in scalability, future efforts may involve exploring approximate kNN search methods, requiring meticulous evaluation of the impact on model performance.

crossbatch Scale: The study reveals the benefits of increasing the crossbatch dimension (d). Current experiments employ values of d = 64 or d = 128, the maximum fitting into a single TPUv3/TPUv2 machine's memory. Future work aims to further elevate d, explore larger memory devices, or adopt multi-node training setups to enhance the scalability of the crossbatch dimension.

Contrastive Learning: FOT training draws inspiration from basic contrastive learning (CL) techniques, contributing to improved key structure and distraction issue mitigation. Future investigations may delve into other CL methods, such as hard negative mining, to harness larger memory effectively during training.

Collaboration with Other Methods: Given the dynamic landscape of long-context methods, the study recognizes the potential for synergies by combining FOT with other emerging techniques. Future research endeavors may explore the integration of FOT with complementary methods, fostering mutually beneficial interactions and advancements in long-context modeling.

\paragraph{Related work.}

The investigation explores diverse methods to extend the contextual range of transformers. Transformer-XL \cite{dai2019transformer}, for instance, caches prior contexts for linear expansion, whereas Longformer \cite{beltagy2020longformer} adopts sparse attention to facilitate token interaction with distant counterparts, thereby reducing computational complexity. Other models such as BigBird \cite{zaheer2020big} and LongT5 \cite{guo2021longt5} similarly employ sparse attention for handling extended sequences. Hierarchical transformers \cite{nawrot2021hierarchical} adopt activation downsampling, and COLT5 \cite{ainslie2023colt5} introduces conditional computation for accommodating larger contexts. Memorizing Transformer \cite{wu2022memorizing} utilizes kNN lookup, aiming to address longer attention context needs and enhance long-context capabilities. Furthermore, the paper delves into the fine-tuning of LLMs for extended retrieval, presenting methods like RETRO \cite{borgeaud2022improving} and Memorizing Transformer \cite{wu2022memorizing}. The proposed methodology extends the model context in a single stage, diverging from retrieval-centric approaches. Additional investigations, such as Landmark attention \cite{mohtashami2023random} and Position Interpolation \cite{chen2023extending, kaiokendev2023things}, focus on extending LLaMA's context length. Notably, the proposed approach eschews reliance on positional encodings, enabling extrapolation to theoretically limitless context lengths. The study also explores zero-shot methodologies, distinguishing from KNN-LM \cite{khandelwal2019generalization} and Parallel Context Windows \cite{ratner2023parallel}. Here, the approach involves fine-tuning models, allowing all tokens to attend to previous tokens within a subset of layers. Finally, the research delves into contrastive learning, setting it apart from CLIP \cite{radford2021learning}, SimCLR \cite{chen2020simple}, TRIME \cite{zhong2022training}, and ContraCLM \cite{jain2022contraclm}. The proposed approach integrates negatives into attention layers, concentrating on training the attention mechanism for extended contexts. It introduces contrastive-inspired techniques tailored explicitly for handling prolonged contexts.

\paragraph{Memory-GPT (MemGPT)} ~

\paragraph{Working.}

MemGPT \cite{packer2023memgpt} introduces a multi-level memory architecture that enables large language models (LLMs) to autonomously manage memory for unbounded context. This architecture distinguishes between two primary memory types: main context and external context. Main context, analogous to a computer's RAM, represents the fixed context window available to the LLM during inference. It comprises three components: read-only system instructions that provide base LLM directives, a read-only FIFO queue storing recent conversational history, and a writable scratchpad for temporary information. Together, these adhere to the processor's maximum context size.

External context, similar to a computer's disk storage, holds information outside the LLM's context window. This out-of-context data can be brought into main context via explicit function calls. External context storage is configurable for specific tasks, like preserving full chat logs for conversational agents or large document collections for analysis.

\begin{figure*}[th]
  \centering
  \includegraphics[width=0.8\linewidth]{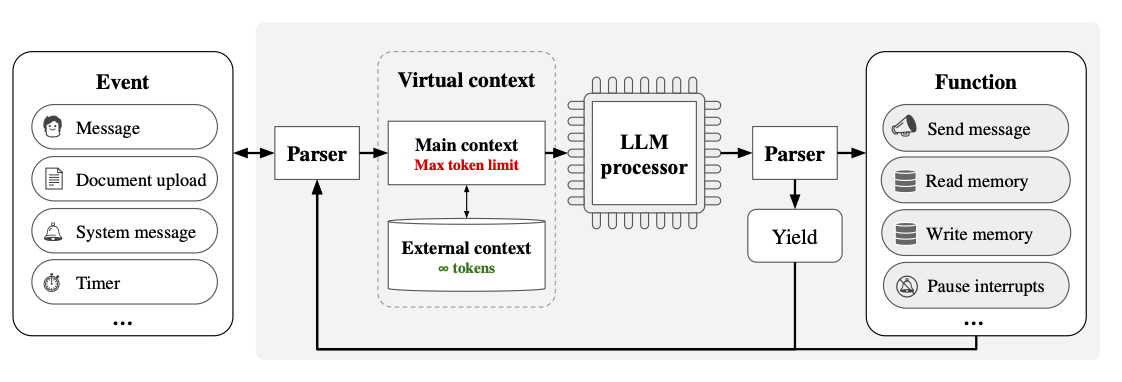}
  \caption{Components of MemGPT \cite{packer2023memgpt}. In MemGPT, a fixed-context language model is enhanced with a hierarchical memory system. The processor manages its memory, using functions to transfer data between main and external contexts. It generates text through a parser, yielding or making function calls, with control requested in advance for chaining functions. The processor pauses during yielding until the next external event.}
  \label{fig:MemGPT}
\end{figure*}

A key innovation in MemGPT is the ability for LLMs to autonomously manage their memory. The pre-prompt provides detailed instructions on the memory hierarchy and utilities, along with a schema of functions to access or modify memory. During each inference cycle, the LLM parses and validates output strings containing memory function calls before execution. This self-directed mechanism is facilitated by a feedback loop enabling the system to learn from its actions. Awareness of token constraints is vital for effective self-editing. MemGPT prompts the LLM processor with warnings about token limits to guide memory decisions.  The control flow in MemGPT is event-triggered, with user messages, alerts, interactions, or timed events initiating inference. Function chaining allows executing multiple functions sequentially, enhancing practical task handling. Functions can return control immediately after completing, adding output to context for continued processing without pausing. Figure \ref{fig:MemGPT} provides an overview of MemGPT's components.

\paragraph{Experiments.}

The research explores the performance of MemGPT in two domains: \textbf{conversational agents} and \textbf{document analysis}. In the conversational agents domain, the study expands the Multi-Session Chat dataset \cite{xu2021beyond}, introducing tasks to assess the agent's knowledge retention and engagement in long conversations. MemGPT is evaluated on criteria of consistency and engagement, showcasing its ability to leverage memory for improved conversation coherence and personalized responses. The study introduces a deep memory retrieval task and evaluates MemGPT against fixed-memory baselines, demonstrating MemGPT's superior performance in maintaining coherence. Additionally, in the conversation opener task, MemGPT exhibits the capability to craft engaging openers by drawing from prior knowledge. In the document analysis domain, the research addresses challenges posed by limited context windows in transformer models. MemGPT is benchmarked against fixed-context baselines in a multi-document question-answering task \cite{liu2023lost}, showcasing its ability to scale effectively to larger context lengths and handle reasoning across documents. The study also introduces a nested key-value retrieval task \cite{liu2023lost}, where MemGPT outperforms GPT-3.5 and GPT-4 by accessing key-value pairs stored in memory, demonstrating its proficiency in multi-hop lookups. The findings highlight MemGPT's effectiveness in both conversational agents and document analysis tasks.

\paragraph{Related work.}

Recent works have focused on improving the ability of large language models (LLMs) to process longer context lengths. This capability is especially useful for conversational agents that require coherent dialogues and for LLMs performing question-answering tasks that need to combine information from multiple sources. Approaches like recursive summarization \cite{wu2021recursively} have been explored to address fixed-length context limitations by generating concise representations over a sliding window. However, this process risks inadvertently losing relevant details.

Given the context length limitations of many LLM applications, there is growing interest \cite{press2021train, dong2023survey, beltagy2020longformer} in enhancing LLMs' capacity for longer sequences. MemGPT can exploit and benefit from expanded context lengths, as it can store more information in its memory. Search and retrieval mechanisms, particularly within the Retrieval-Augmented Generation paradigm, have been integrated into conversational agents for document question-answering, customer support, and chatbots. Various works \cite{lin2023ra, ram2023context, borgeaud2022improving, karpukhin2020dense, lin2023ra, guu2020retrieval} have optimized the retriever or LLM separately, while MemGPT remains agnostic to the specific retrieval method. Recent research has also focused on augmenting LLMs with additional capabilities as interactive agents. Examples include adding memory for planning \cite{park2023generative}, using pagination to control context size in a web environment \cite{park2023generative}, and exploring interleaved reasoning \cite{nakano2112webgpt}. MemGPT specifically tackles equipping agents with long-term memory of user inputs.


\section{Interpolation}
\label{sec:Interpolation}
Interpolation techniques in the context of context length extrapolation focus on fine-tuning or optimizing a model to effectively handle sequences within the range of context lengths it has encountered during training. The emphasis is on refining the model's ability to smoothly extend its comprehension of the context within the observed range, thereby enhancing its performance on sequences within the initially encountered context lengths. These techniques contribute to a more nuanced and improved understanding of context within the trained limits, ensuring that the model performs optimally within the context lengths it has been exposed to during training.

\subsection{Zero-shot extrapolation}
\label{subsec:Zero-shot extrapolation3}
Zero-shot extrapolation for interpolation techniques involves extending a model's capability to handle sequences that fall outside the observed context lengths during training, without explicit fine-tuning or optimization for those lengths. In other words, the model is expected to generalize well to context lengths beyond its training range, relying on the knowledge gained from the observed lengths.

For interpolation, the model is typically fine-tuned or optimized within the observed context lengths. Zero-shot extrapolation, in this context, assesses how well the model performs on longer sequences without any specific adaptation for those lengths. This entails evaluating the model's zero-shot generalization to context lengths that were not explicitly part of its training data.

\subsubsection{Specialized attention mechanism}
\label{subsubsec:Specialized attention mechanism}
In this section, we delve into specialized attention mechanisms designed to address the length generalization failure observed in LLMs when faced with longer contexts. The following papers contribute to this exploration: LM-Infinite \cite{han2023lm}, a solution proposing a $\Lambda$-shaped attention mask and a distance limit for on-the-fly length generalization; LongQLoRA \cite{yang2023longqlora}, an efficient method combining Position Interpolation \cite{chen2023extending}, QLoRA \cite{dettmers2023qlora}, and Shift Short Attention \cite{chen2023longlora} for extending context length with minimal training resources; and LongLoRA \cite{chen2023longlora}, a fine-tuning approach that efficiently extends context sizes while maintaining compatibility with existing techniques. These papers collectively contribute to the advancement of specialized attention mechanisms tailored to mitigate the challenges of zero-shot context length extrapolation in LLMs.

\paragraph{LM-Infinte} ~

\paragraph{Working.}

In the domain of LLMs, a novel approach named LM-Infinite is introduced to address the issue of length generalization in Transformer-based LLMs equipped with relative positional encodings. LM-Infinite presents overarching principles that can be applied across diverse LLMs. LM-Infinite consists of a $\Lambda$-shaped attention mask and a distance limit. The attention mask encompasses global and local branches, allowing tokens to attend to a predefined number of preceding tokens, controlled by a factor denoted as $n_{\text{global}}$. A distance limit is imposed, restricting the \say{effective distance} within the training length limit ($L_{\text{pretrain}}$). LM-Infinite ensures that tokens beyond this limit are excluded during attention, preventing exposure to unseen distances in the pre-training phase. The proposed principles are assessed on three contemporary open-sourced LLM families: LLaMA series (LLaMA and Llama-2), MPT-7B series, and GPT-J series employing various relative positional encoding methods such as RoPE and Alibi encoding. For RoPE \cite{su2021roformer}, involving the rotation of key and query vectors based on positions, LM-Infinite is seamlessly implemented by introducing a global branch with unrotated key vectors and rotated query vectors. In the case of Alibi \cite{press2021train} encoding, offsetting attention logits between tokens, LM-Infinite is integrated smoothly by clipping the offset matrix. This innovative solution offers a promising strategy to overcome challenges related to length generalization in LLMs, augmenting their adaptability to extended contexts during inference. Figure 11 offers an overview of LM-Infinite and a notional model. 

\begin{figure*}[th]
  \centering
  \includegraphics[width=0.68\linewidth]{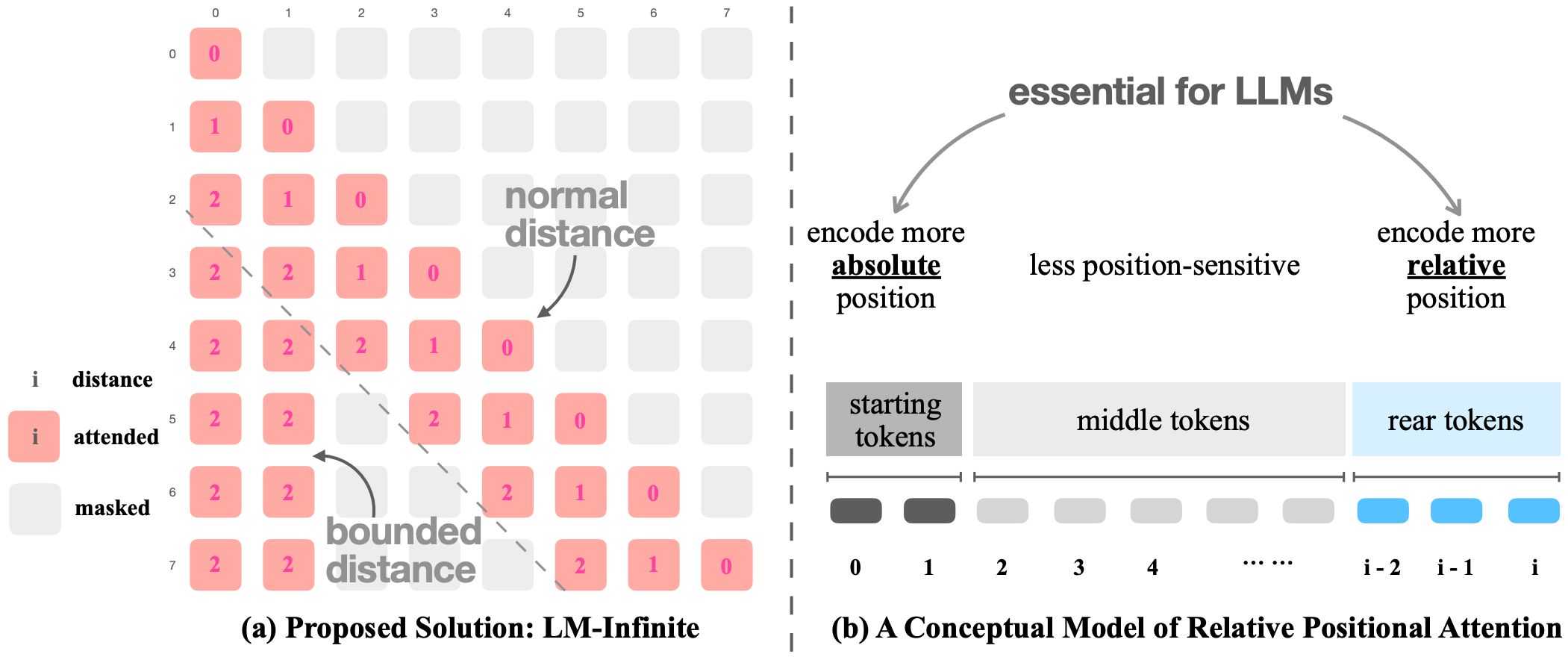}
  \caption{LM-Infinite \cite{han2023lm} is an easy-to-use enhancement for different LLMs, involving a $\Lambda$-shaped mask and a distance constraint during attention. Additionally, a conceptual model to explain how relative position encoding functions is depicted. }
  \label{fig:LM-Infinite}
\end{figure*}

\paragraph{Experiments.}

The evaluation assessed LM-Infinite on the arXiv and OpenWebText2 subsets of the Pile dataset \cite{gao2020pile}, comprising arXiv preprints and Reddit submissions. The fluency was evaluated via perplexity on the arXiv data, demonstrating LM-Infinite successfully flattened the curve for lengths far exceeding training. Consistent fluency in long sequences was observed, with state-of-the-art perplexity scores confirming effectiveness without parameter updates. Notably, MPT-7B+LM-Infinite achieved slightly inferior scores to the fine-tuned MPT-7B-Storywriter, showcasing efficiency as a resource-efficient alternative.

Generation performance was evaluated on arXiv and OpenWebText2 using BLEU \cite{papineni2002bleu} and ROUGE \cite{lin2004rouge} metrics. LM-Infinite extended quality to longer lengths than training, akin to fine-tuning without updates. Analysis revealed diverse effects on different LLMs; LLaMA and GPT-J-6B better maintained quality at longer positions, while Llama-2 performed better at nearer positions. Efficiency assessments demonstrated 3.16x encoding and 2.72x decoding speedups at 32k length. An 8k context example illustrated successful generation.
\paragraph{Diagnosing out-of-distribution (OOD) in LLMs.}

The authors explore out-of-distribution (OOD) factors impacting length generalization challenges in LLMs, using theoretical analysis and experiments. The hypothesis is that while pre-trained LLMs with relative positional encodings can handle relative positions, longer sequences make attention weights and hidden states "unfamiliar," deviating from the training distribution.
\begin{itemize}
\item One key OOD factor is unseen distances. Relative positional encoding relies on attention weights, which may struggle when distances grow beyond anticipated magnitudes. Supported by LLaMA experiments, we present a theorem showing potential explosion of attention logits with increasing length.

\item A second factor is the number of tokens. As texts lengthen, attention weight entropy likely increases, unless logits explode - presenting a tradeoff between the factors. This dilemma is validated theoretically and empirically.

\item Despite lacking absolute position encoding, a third factor shows attention in Transformers with relative positional encodings can implicitly encode it. A theorem and PCA projections demonstrate distinct subspaces for initial tokens.
\end{itemize}

\paragraph{Advantages.}

LM-Infinite presents a groundbreaking approach with the incorporation of two distinctive features: a $\Lambda$-shaped attention mask and the integration of a distance bound during attention. These innovative elements contribute significantly to its appeal, as LM-Infinite eliminates the necessity for parameter updates in pre-trained LLMs, showcasing a remarkable computational efficiency with a complexity of $O(n)$. Beyond this, LM-Infinite demonstrates its practicality through tangible advantages, delivering a substantial 3.16x acceleration in encoding processes and an impressive 2.72x enhancement in decoding efficiency.

\paragraph{Related work.}

Transformer \cite{vaswani2017attention} and its variants, widely utilized in modern LLMs, have gained prominence due to their effectiveness and parallel training capabilities. Positional encodings are crucial for these models, categorized into absolute positional encodings, providing absolute positions using vectors like sinusoidal position embeddings or learned position embeddings, and relative positional encodings, which use distance information between tokens. Examples include learnable attention logit biases in T5 \cite{raffel2020exploring} and Transformer-XL \cite{dai2019transformer}, linear attention decay in Alibi \cite{press2021train}, and techniques like RoPE \cite{su2021roformer}, CAPE \cite{likhomanenko2021cape}, and XPos \cite{sun2022length}.

In the context of fine-tuning on longer texts, existing solutions involve interpolating positional encoding \cite{chen2023extending}, using contrastive learning \cite{tworkowski2023focused}, and adopting padding \cite{tao2023frustratingly} or shifting \cite{kiyono2021shape}. However, these approaches only offer temporary remedies and require substantial training resources. The present work provides an on-the-fly solution by identifying and addressing Out-of-Distribution (OOD) factors affecting length generalization.

Additionally, various efforts have been made to address long-context LLMs. RecurrentGPT \cite{zhou2023recurrentgpt} recurrently generates texts, reading recent context and summaries of longer histories. Some use special mark-up \cite{bueno2022induced} tokens or landmarks \cite{mohtashami2023random} to access informative subsets, while others propose prompting strategies \cite{anil2022exploring} or retrieval-based memories \cite{wu2022memorizing, guu2020retrieval, borgeaud2022improving, khandelwal2019generalization, kaiser2017learning, yogatama2021adaptive}. These designs often necessitate explicit finetuning and lack compatibility with state-of-the-art LLMs. The present work focuses on extending existing LLMs to longer texts dynamically, harnessing their robust generalization capabilities.

\paragraph{LongLoRA} ~

When addressing the processing of exceedingly lengthy sequences, the typical self-attention mechanism \cite{vaswani2017attention} undergoes an escalation in computational expenses, resulting in a deceleration of training and an augmented demand for additional GPU memory. The standard self-attention exhibits a computational complexity of $O(n^2)$, incurring elevated costs on GPU memory. Conversely, shift short attention \cite{chen2023longlora} divides input tokens into clusters and exclusively calculates the attention within each cluster independently. To amplify the information interplay among neighboring clusters, it also calculates attention between the proximate clusters. Employing the sparse local attention mechanism, shift short attention can economize substantial GPU memory. Assuming the input tokens are partitioned into $g$ clusters, the computational complexity can be diminished from $O(n^2)$ to $O((n/g)^2)$.

LongLoRA operates on the premise that, while dense global attention is essential for inference, fine-tuning can be optimally achieved through sparse local attention. The key innovation in LongLoRA involves extending context length during fine-tuning, maintaining a balance between high performance and computational efficiency. This is implemented through an enhanced version of the Low-Rank Adaptation (LoRA) \cite{hu2021lora} method, a well-established technique for streamlining fine-tuning in transformer models. LoRA's distinctive approach involves training and encapsulating additional weight adjustments in a separate matrix, preserving the integrity of pre-trained model weights. This approach streamlines and enhances the efficiency of the fine-tuning process, setting LoRA apart from other methodologies.

\paragraph{Working.}

LongLoRA \cite{chen2023longlora} mitigates the challenge of the computational cost by introducing 2 key aspects: 

\begin{enumerate}
    \item Shift short attention (S2-Attn)

    \item Parameter-efficient fine-tuning
\end{enumerate}

In the fine-tuning phase, S2-Attn employs sparse local attention instead of dense global attention. Essentially, this entails dividing the input document into distinct groups and independently applying attention mechanisms within each group. This segmentation increases perplexity as information exchange between groups is limited. To address this, S2-Attn introduces token shifting by half of the group size, facilitating seamless information exchange between adjacent groups. During this process, the output is coordinately combined, constituting the output of the multi-head self-attention layer, utilizing pre-trained self-attention weights.

\begin{figure*}[th]
  \centering
  \includegraphics[width=0.64\linewidth]{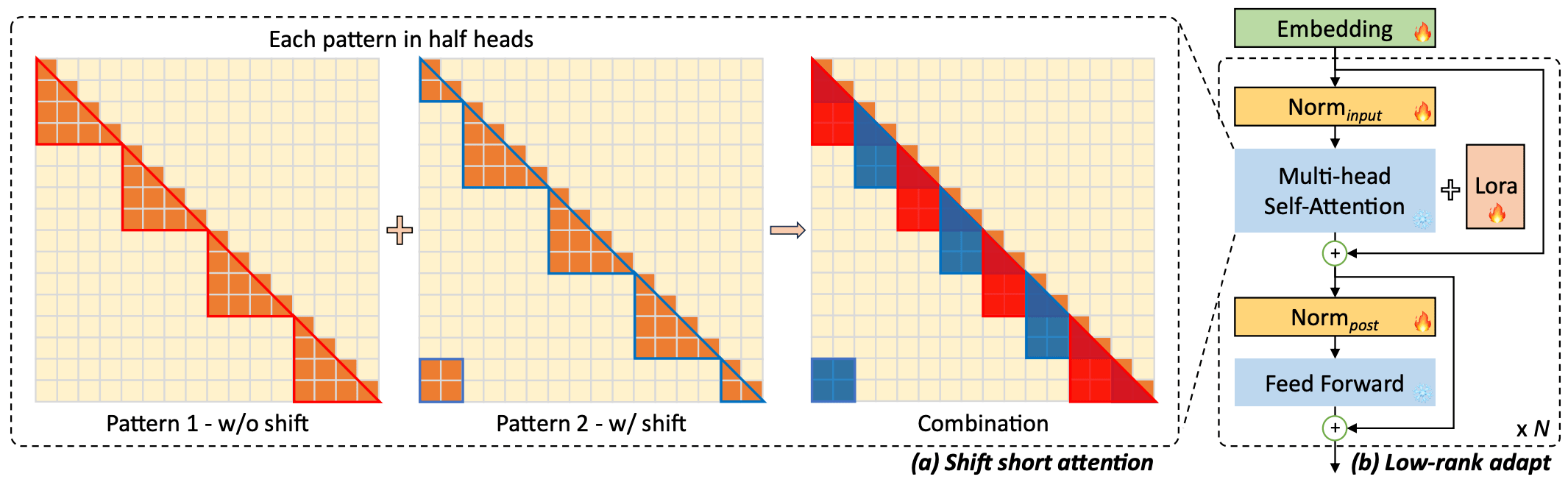}
  \caption{Overview of LongLoRA \cite{chen2023longlora} design. The Shifted Sparse Attention (S2-Attn) is incorporated during fine-tuning, while the trained model maintains its original standard self-attention during inference. LongLoRA extends training by making embedding and normalization layers trainable in addition to LoRA weights in linear layers. This extension is crucial for expanding context, and it introduces only a minimal number of extra trainable parameters. }
  \label{fig:LongLora}
\end{figure*}

\begin{figure*}[th]
  \centering
  \includegraphics[width=0.8\linewidth]{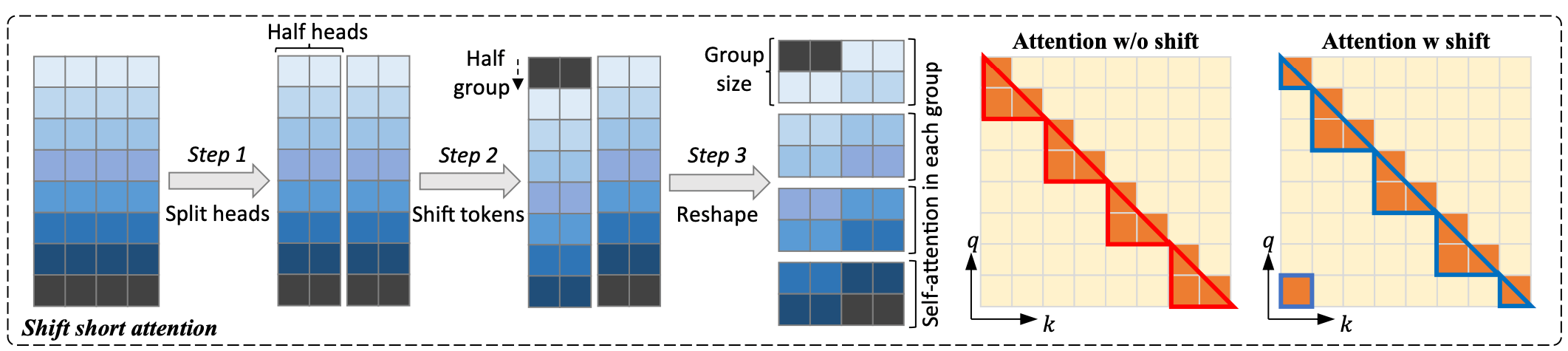}
  \caption{Shift short attention. It involves three steps. Features are split into two chunks along the head dimension. Tokens in one chunk shift by half the group size, and then tokens are grouped and reshaped. Attention is calculated within each group, with information flowing between groups through shifting. \cite{chen2023longlora}}
  \label{fig:Shift short attention}
\end{figure*}

LongLoRA's operational efficiency sees additional enhancements through the reevaluation of the fine-tuning methodology for context expansion. The investigation reveals that integrating LoRA, conventionally employed in attention layers, exhibits significant efficacy when combined with allowing the embedding and normalization layers to learn during the training phase. Figure \ref{fig:LongLora} and Figure \ref{fig:Shift short attention} offer an illustrative explanation of the architecture of LongLoRA and Shift short attention.

\paragraph{Experiments.}

The experimental settings involve extending pre-trained 7B, 13B, and 70B LLaMA2 \cite{touvron2023llama2} models with maximum extended context window sizes ranging up to 100k, 65536, and 32768 for the respective models and utilizing Position Interpolation \cite{chen2023extending}. Training parameters follow Position Interpolation with adaptations for a single 8× A100 GPU machine. Redpajama dataset \cite{together2023redpajama} is used for training, and evaluation is on PG19 \cite{rae2019compressive} and arXiv Math proof-pile \cite{proofpile2022} datasets. A LongQA \cite{chen2023longlora} dataset is created to address chat ability limitations. The main results indicate improved perplexity with longer context sizes. LongLoRA achieves promising results on extremely large settings and retrieval-based evaluations on topic retrieval tasks show comparable performance to LongChat-13B \cite{lymsys2023longchat}, outperforming it in the 16k evaluation.

\paragraph{Advantages.}

LongLoRA has the following advantages:

\begin{itemize}
    \item \textbf{Maintaining the Original Architectural Structure:} S2-Attn fine-tuned models maintain the original attention architecture in inference, enabling seamless integration with established optimization techniques and infrastructure.
    \item \textbf{Integration with Current Techniques and Tools:} LongLoRA seamlessly integrates FlashAttention-2 \cite{dao2023flashattention} and other optimization techniques in both training and inference, facilitating its seamless incorporation into existing workflows.
    \item \textbf{Straightforward implementation:} LongLoRA implementation is straightforward, requiring minimal code for training and optional configuration to retain the original standard self-attention during inference.
\end{itemize}

\paragraph{Related work.}

Numerous studies have delved into extending the context length of transformers. Certain retrieval-based \cite{karpukhin2020dense, guu2020retrieval, izacard2022few} approaches enhanced language models by incorporating related documents into contexts. This work \cite{chen2023longlora}, aligning with such methods, maintains an unaltered attention mechanism during inference. Multiple techniques \cite{zaheer2020big, kitaev2020reformer, qiu2020blockwise, bulatov2022recurrent, beltagy2020longformer, wang2020linformer} approximate multi-head attention to mitigate the quadratic complexity in self-attention computation. Notably, Longformer \cite{beltagy2020longformer} employs sparse attention for handling extended sequences. Others leverage memory mechanisms as compression for past inputs to access relevant tokens. A notable limitation of these techniques is the discernible gap between compression and full attention, hindering the fine-tuning of pre-trained LLMs. Despite entailing an approximation of the attention mechanism, this \cite{chen2023longlora} work preserves a comparable shape and a modest gap to standard attention. This allows for the fine-tuning of pre-trained LLMs while preserving full attention during inference.

\paragraph{LongQLoRA} ~

In this work, \cite{yang2023longqlora} presents LongQLoRA, a memory-efficient and effective method to extend the context length of LLaMA series models. With LongQLoRA, the authors extend the context length of LLaMA2 from 4,096 to 8,192, even to 12k on a single V100 with 32GB memory. LongQLoRA combines the advantages of position interpolation, QLoRA, and shift short attention of LongLoRA.

\paragraph{Working.}

LongQLoRA combines the advantages of Position Interpolation \cite{chen2023extending}, QLoRA \cite{dettmers2023qlora} and Shift Short Attention of LongLoRA \cite{chen2023longlora}. Firstly, it uses Position Interpolation to extend the context length of LLaMA2 \cite{touvron2023llama2} from 4,096 to the target size. To save more GPU memory, during finetuning, it uses QLoRA to quantize the weights of the base model to 4-bit. To further save GPU memory, it also uses Shift Short Attention in finetuning with a group size 1/4 of the target context length.

To recover the performance lost due to imprecise quantization, it adds LoRA \cite{hu2021lora} adapters on all layers, and the LoRA rank is 64. It is found that it achieves better inference performance with standard global attention.

\paragraph{Advantages.}

With a single 32GB V100 GPU, LongQLoRA can extend the context length of LLaMA2 7B and 13B from 4,096 to 8,192 and even to 12k within 1,000 finetuning steps. LongQLoRA achieves competitive perplexity performance on PG19 \cite{rae2019compressive} and Proof-pile \cite{azerbayev2022proofpile} datasets. The model also outperforms LongLoRA and is very close to MPT-7B-8K \cite{team2023introducing} within the evaluation context length of 8,192.

\paragraph{Experiments.}

The study primarily runs experiments on the 7B and 13B models, utilizing a single V100 GPU with 32GB memory throughout the entire experiment. They expand the context length of both LLaMA2-7B and Vicuna-13B models, increasing it from 4096 to 8192.

Initially, Position Interpolation technology is employed to increase the context length from 4096 to 8192. Regarding QLoRA, it quantizes the base model's weights to 4-bit Normal Float \cite{dettmers2023qlora}, sets LoRA rank to 64, and integrates LoRA adapters into all layers.

During the fine-tuning of LLaMA2-7B, they implement the next token prediction task, focusing solely on computing the cross-entropy loss on the target part when fine-tuning Vicuna-13B.

They adopt shift short attention with a group size equivalent to 1/4 of the model’s maximum context length for fine-tuning, utilizing standard global attention during inference.

Regarding the dataset, the authors extract approximately 54k long text samples from the Redpajama dataset \cite{together2023redpajama} to fine-tune pre-trained models, spanning token lengths from 4096 to 32768. Additionally, they conduct perplexity evaluations using the PG19 \cite{rae2019compressive} validation dataset and the Proof-pile \cite{azerbayev2022proofpile} test dataset for pre-trained models.

\paragraph{Related work.}

The LLaMA-series models, like LLaMA and LLaMA2 \cite{touvron2023llama2}, are trained with predetermined context lengths—2,048 for LLaMA and 4,096 for LLaMA2. Their positional encoding, RoPE \cite{su2021roformer}, has limited extrapolation abilities. Once the input length surpasses these preset context lengths, the model's perplexity sharply increases, leading to degraded performance on tasks requiring longer contexts.

Extending the context length by further pretraining demands considerable resources and converges slowly. To address this, techniques like Position Interpolation (PI) \cite{chen2023extending}, focused Transformer (FOT) \cite{tworkowski2023focused}, and LongLoRA \cite{chen2023longlora} have been proposed. However, these methods still require extensive computational resources, often inaccessible to many researchers.

PI \cite{chen2023extending} finetunes LLaMA with 1,000 steps on 32 A100 GPUs to extend the context length from 2,048 to 8,192. FOT \cite{tworkowski2023focused} presents LongLLaMA with 256k context length trained on 128 TPUs. LongLoRA \cite{chen2023longlora} combines PI and LoRA \cite{hu2021lora} to extend LLaMA2’s context length from 4,096 to 100k on 8 A100 GPUs. However, PI and FOT are computationally expensive, and LongLoRA still requires 8 A100 GPUs.

QLoRA \cite{dettmers2023qlora} enables more efficient finetuning by first quantizing models to 4 bits before adding low-rank adapters, reducing memory requirements. This allows finetuning even 65B parameter LLaMA on a single 48GB GPU while matching 16-bit finetuning performance.

\subsubsection{Prompt compression-based approaches}
\label{subsubsec:Prompt compression based approaches}
Prompt compression techniques constitute a pivotal area of exploration within the domain of context length extrapolation for LLMs. As LLMs aim to process longer input sequences or generate extended outputs, the challenge of efficient handling of expansive prompts comes to the forefront. Prompt compression techniques focus on strategies to distill essential information from lengthy prompts while maintaining the integrity and relevance of the input. These methods are designed to enable LLMs to effectively manage extended contexts without sacrificing computational efficiency. In this context, the following paragraph provides an overview of the various prompt compression techniques employed in LLMs, shedding light on their role in enhancing the models' adaptability to diverse input lengths.

\paragraph{LongLLMLingua} ~

\paragraph{Working.}

The LLMLingua framework, as elucidated by \cite{jiang2023llmlingua}, employs a small language model $M_s$ to assess the perplexity of each token within the initial prompt, subsequently eliminating tokens with lower perplexities. The rationale behind this method lies in the notion that tokens with lower perplexities contribute minimally to the overall entropy gain of the language model, making their removal have a negligible impact on the LLM's comprehension. LLMLingua encompasses a budget controller, an iterative token-level prompt compression algorithm, and a distribution alignment mechanism. LongLLMLingua, an extension tailored for long-context scenarios, addresses challenges in enhancing LLM perception of key information relevant to prompt questions. LongLLMLingua delves into four aspects: improving key information density, reducing information loss in the middle, achieving adaptive granular control during compression, and enhancing the integrity of key information.

To enhance key information density, LongLLMLingua introduces both question-aware coarse-grained compression and question-aware fine-grained compression. In coarse-grained compression, it employs a metric $r_k$ to evaluate the importance of each document, aiming to retain documents with higher importance scores. Contrastingly, fine-grained compression assesses the importance of each token in the instruction, question, and retain documents, using contrastive perplexity to represent the association between tokens and questions. This approach aims to ensure that the compressed results contain more question-relevant key information, ultimately improving recall.

Addressing the challenge of information loss in the middle, LongLLMLingua reorders documents based on their importance scores obtained from coarse-grained compression. This strategic reordering aims to optimize LLMs' information perception differences in various positions within the context.

For achieving adaptive granular control during compression, LongLLMLingua dynamically assigns compression budgets based on importance scores from coarse-grained compression. This dynamic allocation ensures that more relevant documents receive a lower compression ratio, allowing for a more nuanced treatment of information based on its relevance to the prompt question.

\begin{figure*}[th]
  \centering
  \includegraphics[width=0.72\linewidth]{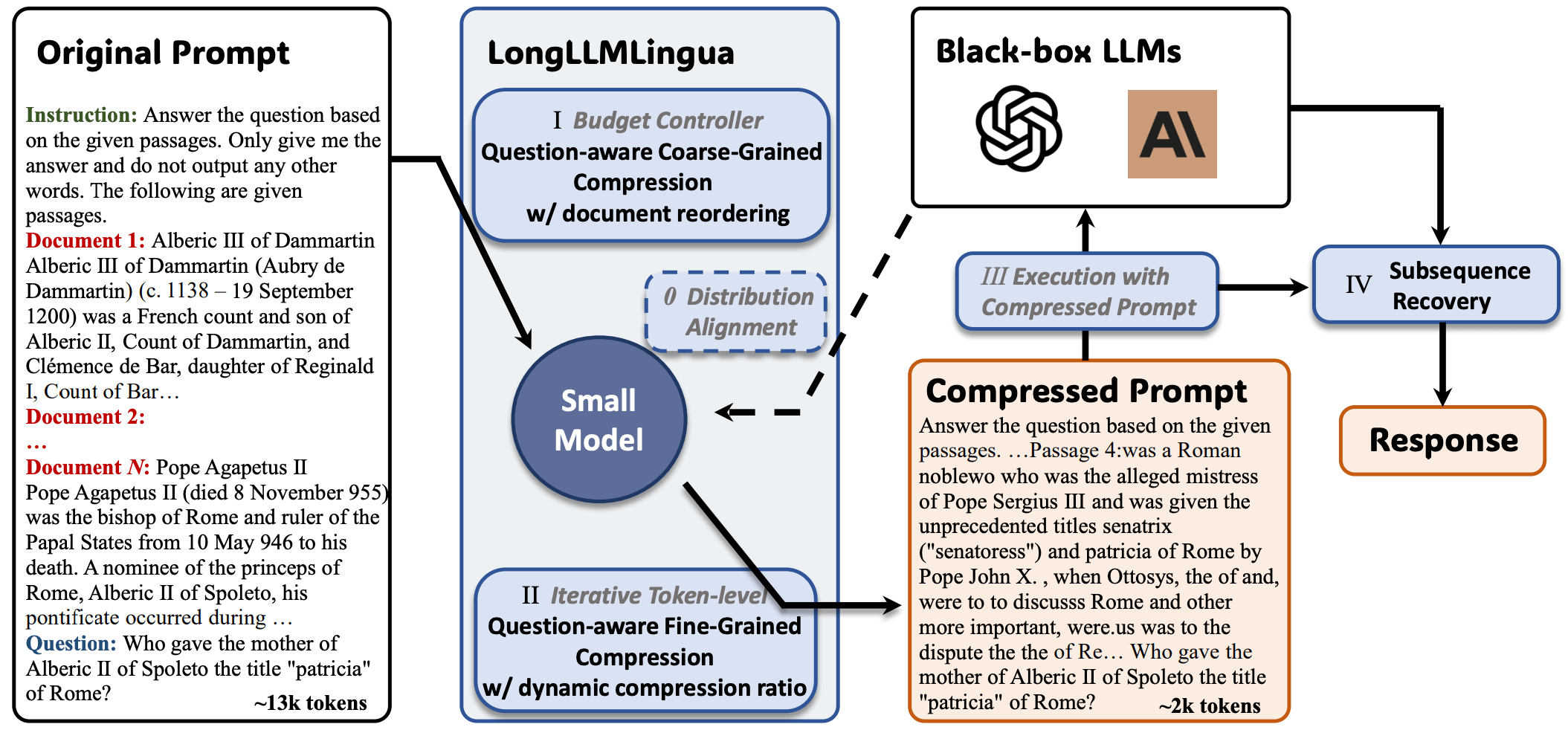}
  \caption{Illustration of LLMLingua's \cite{jiang2023llmlingua} prompt compression: Introducing a budget controller for dynamic compression allocation, applying coarse-grained compression at a demonstration level, detailing an iterative prompt algorithm for knowledge retention, and introducing alignment to address distribution gaps between compact and black-box models.}
  \label{fig:LLMLingua}
\end{figure*}

To enhance the integrity of key information, LongLLMLingua proposes a subsequence recovery method. This method restores the original content from LLMs' responses by iteratively selecting the longest substring that appears in the compressed prompt and mapping it back to the original prompt. This subsequence recovery mechanism aims to rectify potential issues caused by the loss of key information during the compression process, ensuring the accuracy and reliability of information provided to users. Figure \ref{fig:LLMLingua} offers an overview of the LongLLMLingua framework.

\paragraph{Experiments.}

In the experiments section, the research delved into assessing both the effectiveness and efficiency of LongLLMLingua. The chosen LLMs for experimentation were GPT-3.5-Turbo-06134 and LongChat-13B-16k, accessible from OpenAI and HuggingFace. The implementation was carried out using PyTorch 1.13.1 and HuggingFace Transformers, with a focus on stability and reproducibility through the application of greedy decoding and a temperature set to 0. To ensure a consistent basis for comparison, LLaMA-2-7B-Chat was employed for small language models during compression. The datasets selected for evaluation included NaturalQuestions \cite{liu2023lost}, LongBench \cite{bai2023longbench}, and ZeroSCROLLS \cite{shaham2023zeroscrolls}, each serving a distinct purpose in evaluating the performance of LongLLMLingua.

For the NaturalQuestions dataset, which mimics a retrieval-augmented generation setup in commercial search and question-answering scenarios, accuracy served as the primary evaluation metric. LongBench covered a diverse set of tasks, including single-document QA, multi-document QA, summarization, few-shot learning, code completion, and synthetic tasks. The evaluation metrics and scripts provided along with the benchmark were utilized for a thorough assessment. On the other hand, ZeroSCROLLS encompassed summarization, QA, sentiment classification, and reordering tasks across ten datasets.

To establish a baseline for comparison, retrieval-based methods such as BM25, Gzip \cite{jiang2023low}, SentenceBERT \cite{reimers2019sentence}, OpenAI Embedding and the rk metric were employed, along with compression-based methods like Selective Context \cite{li2023unlocking} and LLMLingua \cite{jiang2023llmlingua}. LongLLMLingua consistently outperformed these baselines across a range of tasks and compression ratios, showcasing its effectiveness, particularly in scenarios where irrelevant information was abundant. The proposed document reordering strategy emerged as a valuable enhancement.

A dedicated analysis of latency was conducted using a V100-32G GPU, focusing on the LongBench dataset with an average token count of approximately 10K. The response length was set to 200 tokens in the API call. The results indicated that LongLLMLingua not only facilitated prompt compression but also accelerated the overall inference process. The acceleration effect became more pronounced as the compression rate increased, suggesting its potential significance in scenarios with longer API cost times.

\paragraph{Related work.}

Recent works have explored augmenting the context window of LLMs via strategies like staged pre-training \cite{nijkamp2023xgen}, modifying position embeddings \cite{chen2023extending, peng2023yarn, han2023lm}, implementing linear or sparse attention \cite{ding2307longnet, sun2023retentive}, and incorporating external memory \cite{bertsch2023unlimiformer}. However, their effects on downstream tasks remain unexplored.

Empirical studies show LLM performance diminishes with less effective prompt information \cite{bai2023longbench, lymsys2023longchat, shi2023large}, and depends on the position of pertinent information, with greater challenges comprehending information in the middle versus edges \cite{wu2023selfadaptive, liu2023lost}.

\textls[-10]{Retrieval methods include dense methods using latent vectors like SentenceBERT \cite{reimers2019sentence} and sparse methods based on n-grams like BM25. Recently, \cite{jiang2023low} proposed an unsupervised dense method with compression and kNN.}

Prompt compression methods include: (1) token pruning/merging \cite{goyal2020power, kim2020length, modarressi2022adapler, bolya2022token}; (2) soft prompt tuning like GIST \cite{mu2023learning} and AutoCompressor \cite{chevalier2023adapting}, requiring LLM fine-tuning; and (3) information-entropy techniques like LLMLingua that remove less perplexing tokens.

\subsection{Fine-tuned extrapolation}
\label{subsec:Fine-tuned extrapolation}
Fine-tuned extrapolation for interpolation techniques involves adapting a pre-trained language model to handle longer input sequences not encountered during its initial training. After being primarily trained on sequences within a specified length range (interpolation), the model undergoes a fine-tuning process to improve its performance on longer sequences. This adaptation refines the model's ability to generalize to extended contexts, ensuring seamless handling of both the originally observed and extrapolated input lengths.

\subsubsection{RoPE based approaches}
\label{subsubsec:RoPE based approaches}
\paragraph{Linear positional interpolation} ~

Based on empirical evidence, it has been observed that models undergoing fine-tuning on a pre-existing Transformer with an extended context window exhibit a sluggish adaptation to longer context windows. Position Interpolation emerges as a superior alternative for enabling context window extensions in specific pre-trained LLMs, such as LLaMA when compared to other methods \cite{chen2023extending}. The fundamental concept involves a departure from extrapolation, opting instead to directly scale down the position indices. This scaling ensures that the maximum position index aligns with the prior context window limit during the pre-training phase. In simpler terms, to accommodate a greater number of input tokens, the approach involves interpolating the position encodings at neighboring integer positions. This is a departure from extrapolation beyond the trained positions, which could result in catastrophic values. The theoretical validation of this approach demonstrates that the interpolated attention score possesses a significantly smaller upper bound (approximately 600 times smaller in the LLaMA 7B setting) than its extrapolated counterpart, rendering it more stable. Consequently, the model finds it easier to adapt to interpolated position encodings. Empirically, the effectiveness and efficiency of Position Interpolation are substantiated, requiring only a brief fine-tuning period for the model to seamlessly adjust to substantially extended context windows. The experimental results showcase the extension of the context window from the initial 2048 to 32768 across 7B to 65B LLaMA models through the application of Position Interpolation. 

\begin{figure*}[th]
  \centering
  \includegraphics[width=0.72\linewidth]{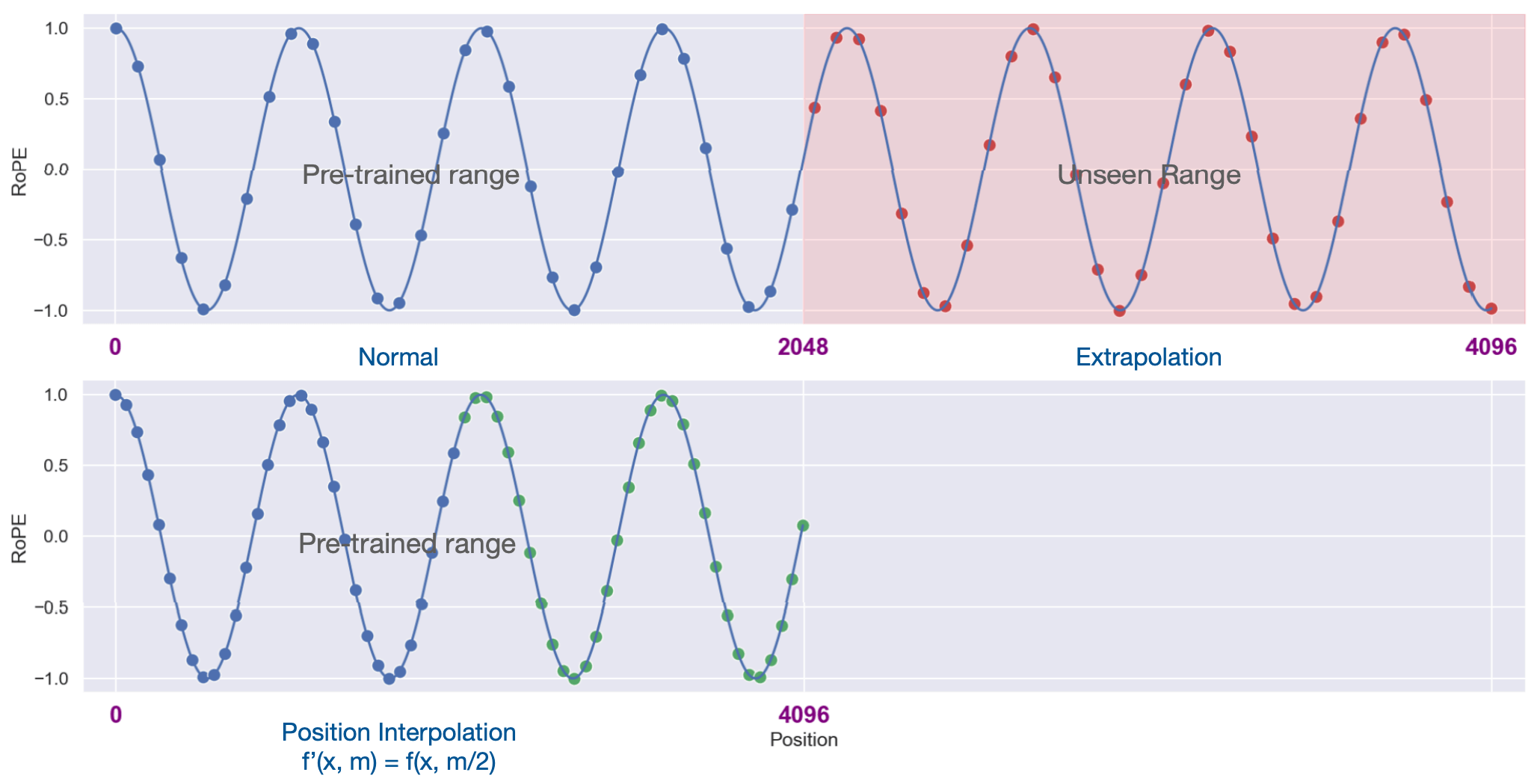}
  \caption{Illustration of Position Interpolation \cite{chen2023extending} method. For a Llama model with a pre-trained context window of 2048 positions, the upper left shows standard usage. In length extrapolation (upper right), the model handles unseen positions (red dots) up to 4096. Position Interpolation (lower left) scales down position indices (blue and green dots) from [0, 4096] to [0, 2048], keeping them within the pre-trained range.}
  \label{fig:Position Interpolation}
\end{figure*}

Figure \ref{fig:Position Interpolation} offers an illustrative explanation of position interpolation.

\paragraph{Experiments and Results.}

In their comprehensive exploration, the researchers showcase the effectiveness of Position Interpolation in extending the context window up to 32 times the original size, achieved with only a few hundred training steps. The resulting models prove to be robust across diverse language tasks, excelling in long language modeling, passkey retrieval, and long document summarization. Variants of the 7B, 13B, 33B, and 65B LLaMA models are extended to various context window sizes up to 32768, employing either direct fine-tuning or the Position Interpolation method. Notably, there are no modifications to the LLaMA model architectures except for rescaling position indices with Position Interpolation. The training procedure involves fine-tuning with the next token prediction objective, utilizing AdamW, linear learning rate warmup, and adjusted batch sizes based on model and context window size. The extended models demonstrate improved perplexity in long-sequence language modeling, with Position Interpolation outperforming direct fine-tuning. Additionally, the models exhibit successful extension of effective context window sizes in a passkey retrieval task. Benchmarks on the original context window size indicate comparable performance with slight regression for longer context windows. In long document summarization, models extended with Position Interpolation achieve competitive ROUGE-1 scores, underscoring their effectiveness in handling this complex task with minimal hyperparameter tuning.

\paragraph{Advantages.}

Utilizing minimal fine-tuning, Position Interpolation emerges as an effective means to considerably expand the context window of LLaMA models. The resulting extended models demonstrate proficiency across a range of tasks within the enlarged context windows, while also preserving their original capabilities for tasks within the initially defined models. This versatility positions them as viable choices for generic language models suitable for both lengthy and concise input prompts. Additionally, models extended through Position Interpolation can leverage existing infrastructure and optimizations, enhancing the practical appeal of this method in various applications.

\paragraph{Related work.}

The research spans several approaches in the field, including retrieval-augmented LLMs. This involves extending LLMs with retrieval modules for incorporating relevant documents into the input context. Notable works in this domain include \cite{karpukhin2020dense, guu2020retrieval, izacard2022few, jiang2022retrieval, khattab2021relevance, santhanam2021colbertv2}. Complementary to these efforts, the study introduces an extended context window for versatility across various tasks like long document summarization, few-shots learning, etc. Another focus is on integrating memory capabilities into Transformers, enhancing their ability to handle lengthy sequences \cite{bulatov2022recurrent, wu2020memformer, dai2019transformer, wu2022memorizing, martins2022inftyformer, mu2023learning}. This work allows attention to all previous tokens, preserving details without compression. Unlike landmark attention \cite{mohtashami2023random, chen2023extending}, it enables full access to the entire input through unmodified attention. The study is also compatible with Approximated Multi-head Attention methods, reducing the memory and computational complexity of the multi-head attention mechanism through approximation and sparsification \cite{child1904generating,zaheer2020big, beltagy2020longformer, wang2020linformer, choromanski2020rethinking, kitaev2020reformer, ren2021combiner}. In the realm of length extrapolation, recent research \cite{press2021train, sun2022length, haviv2022transformer} is dedicated to training Transformers on short sequences for subsequent inference on lengthier ones. However, these approaches have yet to be implemented in prominent language models like LLaMA \cite{touvron2023llama} or OPT \cite{zhang2022opt}, limiting their applicability to well-established pre-trained models. \cite{chen2023extending} addresses this gap by extending existing LLMs, providing a cost-effective solution for length extrapolation while maintaining the original models' quality, even in tasks with modest context windows. \cite{dosovitskiy2020image} introduced linear interpolation of learned position embeddings in Vision Transformers to support higher resolution during fine-tuning. In contrast, the current work interpolates position indices, tailored for RoPE-like position encodings, potentially requiring less training with no additional trainable parameters. Achieving successful context window extension up to 32 times, \cite{chen2023extending} surpasses \cite{dosovitskiy2020image}'s exploration of up to 4 times, confirming the effectiveness of Position Interpolation for extending context windows in language models.

\paragraph{Yet another RoPE extensioN (YaRN)} ~

This paper \cite{peng2023yarn} addresses a persistent limitation in positional encodings related to their inability to generalize beyond the context window seen during training. While some methods like ALiBi show limited generalization, none can extend to significantly longer sequences than their pre-trained length. Previous works proposed solutions like Position Interpolation \cite{chen2023extending} and "NTK-aware" interpolation \cite{ntkaware2023}, with applications in open-source models like Code Llama \cite{roziere2023code} and Qwen 7B \cite{bai2023qwenvl}. In this paper, YaRN, an enhanced extension method for RoPE in models like LLaMA, GPT NeoX, and PaLM is introduced. YaRN achieves state-of-the-art performance in extending context windows, requiring fine-tuning on less than 0.1\% of the original pre-training data. Additionally, Dynamic-YaRN, coupled with the Dynamic Scaling inference-time technique, achieves over 2x context window extension without fine-tuning. 

\paragraph{Structure of YaRN.}

The paper investigates the challenges associated with Position Interpolation falling short in predicting the complex relationship between RoPE and LLM dynamics. The "NTK-aware" interpolation is introduced to address the loss of high-frequency details but suffers from sub-optimal extrapolation.

To overcome these limitations, the paper presents the "NTK-by-parts" interpolation, targeting specific RoPE dimensions to improve performance. It explores Dynamic Scaling, specifically "Dynamic NTK" interpolation, which gracefully handles repeated forward passes and demonstrates effectiveness, especially on non-fine-tuned models.

In the YaRN method, incorporating temperature reparametrization on logits to modify the attention mechanism without altering the code is proposed. YaRN proves efficient in context extension with zero overhead during both training and inference. The combination of YaRN with "NTK-by-parts" interpolation further enhances the proposed method.

During the training phase, the models were expanded without altering their architecture. The learning process involved recalculating certain aspects with different values. The training was conducted in steps, adjusting parameters like the learning rate and employing specific techniques. This process was performed for two model sizes, namely 7B and 13B parameters, and specific strategies were applied to optimize their performance.

In the extrapolation and transfer learning phase, the models were evaluated on their ability to apply what they learned to new scenarios. This involved using a dataset with a certain context length, and the models were fine-tuned to enhance their capabilities. The evaluation included exploring how well the models could adapt to new, longer contexts, exceeding what they encountered during the original training.

The results indicate that the larger model successfully adapted to extended context lengths, showcasing its ability to learn and apply knowledge from different scales. This adaptation was achieved efficiently, demonstrating the model's effectiveness in transferring its learned information to new contexts without the need for extensive relearning.

\paragraph{Results.}

The evaluation commenced by scrutinizing the model's performance with an expanding context window. To achieve this, ten random samples from Proof-pile, each containing at least 128k tokens, were selected. The perplexity of these samples was assessed across different sequence lengths, ranging from 2k to 128k tokens, with truncation at 2k steps. It's noteworthy that the training methodologies for PI and "NTK-aware" models followed \cite{chen2023extending}, while YaRN used a similar approach with 2.5x fewer training steps and data.

The findings indicate robust performance throughout the targeted context sizes, with YaRN interpolation notably extending the effective context size of Llama-2 to 128k. Particularly noteworthy are the YaRN (s=32) models, exhibiting a continuous decline in perplexity through 128k despite fine-tuning limited to a 64k token length. This demonstrates the model's ability to generalize to unseen context lengths.

The passkey retrieval task, defined in \cite{mohtashami2023landmark}, assesses a model's capability to retrieve a simple passkey from a large amount of otherwise meaningless text. The evaluation involved ten iterations of the passkey retrieval task with the passkey placed at random locations uniformly distributed across evaluation context windows of different sizes, ranging from 8k to 128k. Both 7b and 13b models, fine-tuned using YaRN at a 128k context size, demonstrated very high accuracy (>99\%) across the entire context window size.

Minimal performance degradation was observed between the YaRN models and their respective Llama-2 baselines. On average, there was a mere 0.49\% drop in scores between the YaRN (s=16) and YaRN (s=32) models. This suggests that the iterative extension from 64k to 128k results in negligible performance loss.

In summary, YaRN represents an improvement over existing RoPE interpolation methods and can seamlessly replace Position Interpolation without drawbacks and with minimal implementation effort. The fine-tuned models maintain their original capabilities across multiple benchmarks while effectively attending to significantly larger context sizes. Additionally, YaRN facilitates efficient extrapolation through fine-tuning on shorter datasets and leverages transfer learning for faster convergence, addressing essential aspects in compute-constrained scenarios.

\paragraph{Related work.}

ReRoPE \cite{rerope2023} seeks to expand the context size of RoPE-based pre-trained models, claiming an \say{infinite} context length without fine-tuning, supported by a monotonically decreasing loss up to 16k on the Llama 2 13B model. Unlike embedding interpolation methods, ReRoPE achieves context extension by modifying the attention mechanism. However, it is currently incompatible with Flash Attention 2 \cite{dao2023flashattention} and necessitates two attention passes during inference. LM-Infinite, a concurrent proposal, shares similar concepts with YaRN but emphasizes \say{on-the-fly} length generalization for non-fine-tuned models. Like ReRoPE, LM-Infinite modifies the attention mechanism, making it incompatible with Flash Attention 2 and not a direct embedding interpolation method.

\paragraph{Positional Skip-wisE (PoSE)} ~

\cite{zhu2023pose} present a method called Positional Skip-wisE (PoSE) training to effectively extend the context window for transformers. They split the fixed-length context window into multiple chunks and introduce specialized position bias terms for each chunk. By varying the bias values and lengths of the chunks on a per-example basis, the model learns representations for sequence positions across a longer range than the actual context size. In this way, PoSE training can simulate having a larger context window using only a fixed window, enabling the model to accumulate long-range contextual knowledge. 

\paragraph{Working.}

The PoSE technique simulates longer inputs by manipulating position indices within a fixed context window. It partitions the original window into chunks, adjusting the position indices of each chunk by adding distinct skipping bias terms. These terms, along with the chunk lengths, vary for each training example, allowing the model to adapt to all positions within the target context window during fine-tuning. By maintaining continuous position indices within each chunk, PoSE closely resembles pre-training, retaining the model's language modeling and comprehension abilities.

To illustrate, the method divides the original context window $L_c$ into $N$ chunks $c_0, c_1, \ldots, c_{N-1}$, each with lengths $l_0, l_1, \ldots, l_{N-1}$, where $\sum_{i=0}^{N-1} l_i = L_c$. It introduces the starting index $st_i$ for each chunk $c_i$, facilitating the formulation of position indices as follows:
\begin{equation}
\text{Pos}(c_i) = \{st_i, st_i + 1, \ldots, st_i + l_i - 1\},
\end{equation}
where,
\begin{equation}
\quad st_i = \sum_{j=0}^{i-1} l_j
\end{equation}

Subsequently, it employs the discrete uniform distribution $U(S)$ to sample a skipping bias term $u_i \sim U(\{u_{i-1}, \ldots, L_t - L_c\})$ for each chunk $c_i$.

This bias term transforms the original position indices into:
\begin{equation}\noindent
 \Scale[0.75]{\text{PoSE}(c_i) = \{ui + st_i, ui + st_i + 1, \ldots , ui + st_i + l_i - 1\}}
\end{equation}

Note that the constraint $ui \geq ui - 1$ prevents position index overlaps between chunks.

For the text within each chunk, a similar procedure selects continuous spans of tokens from the input text $x = \{x_0, x_1, \ldots, x_{L_x}\}$. 

After settling position indices and content for each chunk, it performs position interpolation to stabilize fine-tuning.

\begin{figure*}[th]
  \centering
  \includegraphics[width=0.72\linewidth]{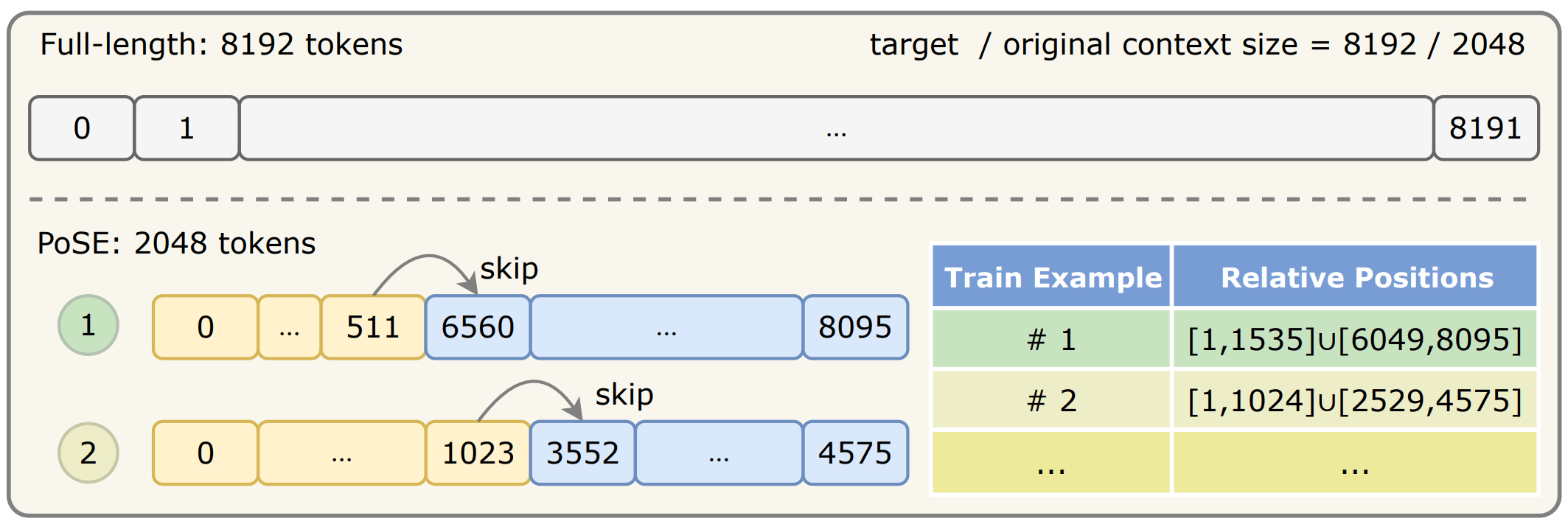}
  \caption{Illustration of Full-length fine-tuning vs PoSE \cite{zhu2023pose} fine-tuning for extending the context from 2,048 to 8,192 tokens. Full-length uses all 8,192 tokens directly, while PoSE adjusts the position indices of 2,048 tokens with a unique skipping bias term. This allows the model to adapt to different relative positions during fine-tuning. }
  \label{fig:PoSE}
\end{figure*}

Figure \ref{fig:PoSE} offers an illustrative comparison of full-length fine-tuning and PoSE.

\paragraph{Advantages.}

The advantages of PoSE are threefold:

\begin{enumerate}
    \item \textbf{Memory and Time Efficiency}: By necessitating only the original context size for fine-tuning, PoSE avoids the quadratic increase in computational complexity concerning the target length during the fine-tuning stage. This significantly diminishes memory and time overhead.

    \item \textbf{Potential for Extremely-Long Context}: The PoSE architecture demonstrated success in extending the context window of LLaMA \cite{touvron2023llama}, magnifying the context window from an original 2,048 tokens up to 131,072 tokens - a 64 times gain. Despite this major expansion in context length, PoSE maintained LLaMA's capabilities for language modeling and understanding.

    \item \textbf{Compatibility with RoPE-based LLMs and PI Strategies}: The effectiveness of PoSE has been empirically validated across several representative RoPE-based LLMs, including LLaMA, LLaMA2 \cite{touvron2023llama}, GPT-J \cite{wang2021gpt}, and Baichuan \cite{yang2023baichuan}. Furthermore, PoSE has demonstrated compatibility with various position interpolation methods, including Linear \cite{chen2023extending}, NTK \cite{ntkaware2023}, and YaRN \cite{peng2023yarn} interpolation.
\end{enumerate}

\paragraph{Experiments.}

The authors investigate the efficacy of long-text modeling across two primary tasks: language modeling and passkey retrieval. Language modeling serves as a fundamental measure of a model's overall proficiency in handling extensive text, while passkey retrieval gauges the maximum token distance considered during the inference stage. Language modeling is evaluated on the GovReport dataset \cite{huang2021efficient} and the Proof-pile dataset \cite{azerbayev2022proofpile}. For passkey retrieval, they adopt the approach outlined in \cite{mohtashami2023landmark} to generate synthetic prompts for assessment.

Experiments were conducted by scaling to 16k and 32k using Full-length training on two techniques: RandPos \cite{ruoss2023randomized} and PoSE. Perplexity scores at various evaluation context window sizes (ranging from 2k to 32k) are reported for each scaled model, including the non-fine-tuned LLaMA model (None). The sliding window approach proposed by \cite{press-etal-2021-shortformer} is employed for evaluation, with a window stride set to 1,024 for efficiency.

Observations include an overall decreasing perplexity trend for both 16k- and 32k-scaled models via PoSE. Despite significantly shorter context lengths during fine-tuning, PoSE achieves comparable results with Full-length, affirming its effectiveness. Notably, PoSE outperforms RandPos.

In the passkey retrieval test proposed by \cite{mohtashami2023landmark}, models are tasked with recovering a random passkey concealed within an extensive document. The none-fine-tuned LLaMA model (None) experiences a rapid drop in retrieval accuracy to 0 when the prompt length exceeds 2k. In contrast, both PoSE-extended models maintain a high retrieval accuracy ($\geq 90\%$) within their respective target context windows. This suggests that models trained via PoSE genuinely possess the capability to attend to all tokens within the extended context windows.

\paragraph{Related work.}

Several models, including those proposed by \cite{press2021train, sun2022length}, and \cite{haviv2022transformer}, aim to ensure consistent performance even when the number of input tokens during inference surpasses the model's trained context window size. On the other hand, some works \cite{chen2023extending, ntkaware2023, peng2023yarn, chen2023longlora} have focused on fine-tuning LLMs with a longer context window. However, all these methods require Full-length fine-tuning,
suffering computational cost that grows with the target context size.
\cite{ruoss-etal-2023-randomized} also attempted to mimic longer sequences during training to address out-of-distribution lengths. They introduced randomized positional encoding (RandPos), randomly selecting an ordered subset of position indices from longer sequences.  PoSE differs significantly from RandPos: RandPos primarily enhances the length generalization abilities of encoder-only models during pre-training. In contrast, PoSE efficiently extends the context window of pre-trained LLMs, especially those with a decoder-only architecture. Furthermore, while RandPos lacks continuous position indices between adjacent tokens, PoSE deliberately maintains this continuity within each chunk. This continuity closely aligns with the pre-training phase, minimizing any disruption to learned language modeling and comprehension abilities.

Managing extremely long input sequences often involves memory mechanisms. Two prominent research lines leverage memory: the recurrence-based approach (\cite{dai2019transformer, bulatov2022recurrent}) and the retrieval-based strategy (\cite{wu2022memorizing, wang2023augmenting, tworkowski2023focused}). Recurrence-based methods segment lengthy inputs, reusing hidden states from previous segments as memory for the current one. However, they suffer from information loss and limited random access capacity. In contrast, retrieval-based paradigms encode prior sequences as (key, value) pairs, using a memory retriever and reader to extract encoded information. A drawback here is the lack of interaction between discrete memory segments.

Recently, \cite{mohtashami2023landmark} proposed landmark attention, enabling random access to input chunks via landmark tokens. However, the PoSE method achieves full access to the entire input without altering the attention mechanism.

\section{Conclusion}

In summary, this paper provides a comprehensive review of the diverse techniques and methodologies for extending the context length of LLMs. The taxonomy presented categorizes these approaches into two broad strategies - \textbf{extrapolation} and \textbf{interpolation}. Extrapolation techniques aim to expand the model's ability to handle sequences beyond its initially trained context length. This includes zero-shot methods leveraging specialized components like position encodings, attention mechanisms, and memory augmentation to achieve on-the-fly generalization. Fine-tuning strategies are also explored to adapt models for longer contexts not encountered during pre-training. Interpolation techniques focus on optimizing models to smoothly extend context comprehension within the observed training length. Specialized attention mechanisms and prompt compression facilitate efficient handling of lengthy contexts. Fine-tuning interpolation adapts models to gracefully transition when sequences begin to exceed the trained length. The survey offers insights into the versatility of techniques spanning prompt engineering, attention mechanisms, positional encodings, and memory augmentation. It highlights innovations in model architectures and training methodologies tailored to address context length limitations. Extensive empirical analysis substantiates the efficacy of these diverse techniques on benchmarks and downstream tasks. By providing a structured taxonomy and review of existing literature, this paper contributes to a clearer understanding of the evolving landscape of context length extension in LLMs. The discussions identify promising research directions, underscoring the importance of continued efforts to develop models proficient in processing extensive contextual information. With increasing interest in long-form text generation and reasoning over large corpora, enhanced context handling will remain an active area of research in coming years.

\section{Discussion}

This comprehensive survey highlights the remarkable progress made in developing diverse methodologies for expanding the contextual capacities of LLMs. However, several open questions and challenges remain, warranting further investigation by the research community.
A key direction for future work involves exploring synergistic combinations of the techniques reviewed in this paper. For instance, integrating memory augmentation strategies with specialized attention mechanisms could potentially yield models proficient in handling significantly longer contexts. Hybrid approaches that leverage the complementary strengths of different techniques merit deeper exploration.
Another crucial aspect requiring attention is the development of appropriate evaluation benchmarks and metrics for accurately assessing context extension techniques. While preliminary benchmarks have been proposed, standardized suites could facilitate more rigorous comparisons between methods. Metrics that provide nuanced insights into a model's contextual capacities beyond simplistic perplexity scores will be valuable.
The interpretability of context extension techniques is also an underexplored area. Methods that enhance the explainability of how models utilize extended contexts could unlock deeper insights into their inner workings. This interpretability will be key for debugging, analysis and responsible deployment of LLMs.
Training efficiency and the high resource costs of developing models with expanded contexts is a significant challenge. Techniques that can match the efficiencies of training natively on short contexts could accelerate progress. Multi-stage training procedures and transfer learning for context extension are promising directions.
Finally, studying the impact of long contexts on emergent capabilities of LLMs presents intriguing opportunities. For instance, how does reasoning over documents rather than sentences transform a model's understanding of complex concepts? Investigating these higher-order effects through carefully designed evaluations and experiments remains an open avenue for future work.
In conclusion, this survey provides a structured foundation that summarizes progress and outlines key open challenges. Continued research leveraging this synthesis of existing literature will further the development of LLMs that exhibit an intricate awareness of long-range context. With higher contextual sophistication, these models are poised to ultimately attain more human-like language comprehension.






\bibliography{custom}

\appendix




\captionsetup{width=16cm}
\tiny
\clearpage
\onecolumn
\renewcommand{\arraystretch}{1.8}
\begin{longtable}
{|P{0.45in}|P{0.4in}|P{0.8in}|P{0.46in}|P{0.5in}|P{0.6in}|P{0.5in}|P{0.5in}|P{0.55in}|}
    \caption{Summary of all the works related to context length extension techniques. Here, we have divided each work by the following factors: 1. Technique, 2. Train Length(s), 3. Evaluation Length(s), 4. Metrics, 5. Model(s), 6. Task(s) and 7. Benchmark(s).}\label{tab:summary} \\ 
    \hline 
    \raggedbottom \textbf{Category}&
    \textbf{Technique}&
    \textbf{Title}&
    \textbf{Train Length(s)}&
    \textbf{Evaluation Length(s)}&
    \textbf{Metric(s)}&
    \textbf{Model(s)}&
    \textbf{Task(s)}&
    \textbf{Benchmark(s)}\\
    \hline
    \endfirsthead
    \multicolumn{9}{c}%
    {\tablename\ \thetable\ -- \textit{Continued from the previous page}} \\
    \hline
    \textbf{Category} &  \textbf{Technique}    &   \textbf{Title} &   \textbf{Train Length(s)} & \textbf{Evaluation Length(s)} & \textbf{Metric(s)} & \textbf{Model(s)}& \textbf{Task(s)} & \textbf{Benchmark(s)}\\
    \hline
    \endhead
    \hline \multicolumn{5}{r}{\textit{Continued on the next page}} \\
    \endfoot 
    \hline
    \endlastfoot

\multirow{14}{*}[-30ex]\textbf{Extrapolation:} \textbf{Zero-shot} &
\centering Positional encoding &
\raggedright Train Short, Test Long: Attention with Linear Biases Enables Input Length Extrapolation \cite{press2021train} &
64, 128,\newline 256, 512, 1024, 1536, 2048, 3072&
\raggedright 64, 128, 256, 512, 1024, 1536, 2048, 3072&
\raggedright Perplexity, Words per seconds, Memory & Customized Transformer Model & Language modeling, text generation 
& WikiText-103,  Toronto  Book Corpus, CC100+RoBerta Corpus
\\
\cline{3-9}
& & \raggedright RoFormer: Enhanced transformer with Rotary Position Embedding \cite{su2021roformer} & 128, 256, 512, 1536 & Details not provided & BLEU, GLEU, Accuracy & Customized Transformer Model (RoFormer), BERT, WoBERT,  NEZHA & Machine translation, Semantic text matching
& WMT 2014 English-German dataset, Chinese 
dataset, CAIL2019-SCM, GLUE( MRPC,  SST-2, QNLI,  STS-B, QQP,  MNLI), Wikipedia Corpus Foundation, BookCorpus

\\
\cline{3-9}
& & \raggedright Randomized Positional Encodings Boost Length Generalization of Transformers \cite{ruoss2023randomized} & 1024, 2048, 4096, 8192 & Details not provided & Accuracy & Encoder-only Customized Transformer Model & Various algorithmic reasoning tasks
& Manual

\\

\cline{2-9}

&
\centering Specialized attention mechanism
&
\raggedright A Length-Extrapolatable Transformer \cite{sun-etal-2023-length} & 1024 & 256, 512, 1024, 2048, 4096 & Perplexity & Customized Transformer Model (LeX) & Details not provided & arXiv dataset, Pile, Books3, 
OpenWebText2, Stack Exchange, 
PubMed Abstracts, Wikipedia, PG-19, 
BookCorpus2, NIH Exporter, Pile-CC\\
\cline{3-9}

& & \raggedright LongNet: Scaling Transformers to 1,000,000,000 Tokens \cite{ding2023longnet} & 8k, 16k, 32k &
2k, 8k, 32k
& Perplexity & Customized Transformer Model (LongNet) & Long-sequence modeling
& Stack dataset \\

\cline{2-9}

&
\centering Window based approaches &
\raggedright GrowLength: Accelerating LLMs Pretraining by Progressively Growing Training Length \cite{jin2023growlength}  & 128, 256, 512, 1024 & Details not provided & \raggedright Training loss & 70M, 160M and 410M LLM (specific model not mentioned) & Details not provided & Neural networks and
the chomsky hierarchy \\

\cline{2-9}

&
\centering Memory/Retri-eval augmented approaches &
\raggedright Landmark Attention: Random-Access Infinite context length for Transformers \cite{mohtashami2023landmark}  & 250, 256, 300, 360, 512 & 512, 2048, 4096 & \raggedright Perplexity & Transformer-XL, LLaMA-7B & Language modeling, Next word prediction, Information retrieval over long contexts & PG-19, arXiv \\
\cline{3-9}

& & \raggedright Augmenting Language Models with Long-Term Memory
 \cite{wang2023augmenting} & 1k & 65k
& Perplexity, accuracy, F1 score & GPT-2-407M & Long-context language modeling, long-context understanding, memory-augmented in-context learning
& Gutenberg-2022 (PG-22), arXiv, ChapterBreak, SST-2, MPQA, MR, Subj, SST-5 \\

\cline{1-9}
\multirow{14}{*}[-20ex]\textbf{Extrapolation:} \textbf{Fine-tuned} &
\centering Memory/Retri-eval augmented approaches&
\raggedright  Think-in-Memory: Recalling and Post-thinking Enable LLMs with Long-Term Memory \cite{liu2023thinkinmemory}&
Details not provided &
\raggedright Details not provided &
Retrieval Accuracy, Response Correctness, Contextual Coherence &
\raggedright ChatGLM-6B, Baichuan2-13B &Response generation for long-term conversations & KdConv, GVD, RMD\\

\cline{3-9}

& &
\raggedright Focused Transformer: Contrastive Training for Context Scaling \cite{tworkowski2023focused}  & 512, 1024, 2048, 4096 & \raggedright 2k, 4k, 8k, 16k, 64k, 128k & Perplexity & Decoder-only Customized Transformer Model (LongLLaMA), OpenLLaMA 3B, OpenLLaMA 7B & Passkey retrieval, QA, Long-context language modeling & PG-19, arXiv, Github code, Isabelle, TREC, WebQS \\
\cline{3-9}

& & \raggedright MemGPT: Towards LLMs as Operating Systems \cite{packer2023memgpt} & Details not provided &Details not provided & ROUGE-L, Accuracy, CSIM similarity scores & GPT-4 & Deep memory retrieval task, Conversation opener task, Multi-document QA, Nested key-value retrieval requiring multi-hop lookups &  
MSC, NQ-Open Wikipedia \\

\cline{1-9}

\multirow{14}{*}[-20ex]\textbf{Interpolation}: \newline \textbf{Zero-shot} &
\centering Specialized attention
mechanism&
\raggedright  LM-Infinite: Simple On-the-Fly Length Generalization for Large Language Models \cite{han2023lminfinite}&
Details not provided &
\raggedright 2k, 4k, 8k, 16k, 32k, 128k &
\raggedright Perplexity & MPT-7B, LLaMA, GPT-J-6B, LLaMA-2
& Long text generation & arXiv, OpenWebText2\\
\cline{3-9}
& & \raggedright LongLoRA: Efficient Fine-tuning of Long-Context Large Language Models \cite{chen2023longlora} & 2048, 4096
&
8192, 16384, 32768, 65536, 100,000

& Perplexity,
Passkey retrieval accuracy,
Win-rate in topic retrieval
 & LLaMA2-7B, 13B, and 70B & Long-sequence language modeling, Topic retrieval & RedPajama
\\
\cline{3-9}
& & \raggedright LongQLoRA: Efficient and Effective Method to Extend context length of Large Language Models \cite{yang2023longqlora}  & 4k, 8k
& 1024, 2048, 
4096, 8192 & \raggedright Perplexity & LLaMA-2-7B, MPT-7B, 
LongLoRA-7B, LongQLoRA & Long-sequence language modeling,
Passkey retrieval,
Topic retrieval in long conversations & PG-19, Proof-Pile \\

\cline{2-9}

&
\centering Prompt compression based
approaches&
\raggedright LongLLMLingua: Accelerating and Enhancing LLMs in Long Context Scenarios via Prompt Compression \cite{jiang2023longllmlingua} & 3k-10k 
&

Details not provided
& Accuracy & GPT-3.5-Turbo, 
LongChat-13B & Single and Multi-Doc QA,
Summarization,
Code completion,
Sentiment classification,
Information reordering & NQ-multi-document QA, LongBench, ZeroSCROLLS \\
\cline{1-9}

\multirow{14}{*}[-20ex] \textbf{Interpolation:} \newline \textbf{Fine-tuned} &
\centering RoPE based approaches &
\raggedright  Extending Context Window of Large Language Models via Positional Interpolation \cite{chen2023extending}&
8192, 
16384, 32768
&
\raggedright 2048, 4096, 
16384, 32768 &
\raggedright Perplexity & LLaMA-7B, 13B, 33B, and 65B & Language modeling, Passkey retrieval, Long document summarization & Pile, PG-19, RedPajama, arXiv math \\
\cline{3-9}
& & \raggedright YaRN: Efficient Context Window Extension of Large Language Models \cite{peng2023yarn} & 4k, 32k, 64k, 
100k, 128k &

8192, 32768, 
65536, 98304, 
131072 
& Perplexity & LLaMA-2 7B, LLaMA-2 13B, GPT-NeoX and Mistral 7B v0.1,MistralLite 7B, PaLM & Passkey retrieval & PG-19, Proof-pile, Hugging Face open LLM benchmark suite
\\

\cline{3-9}
& & \raggedright PoSE: Efficient Context Window Extension of LLMs via Positional Skip-wise Training \cite{zhu2023pose} & 2k, 16k &

2k, 4k, 8k, 16k, 
32k, 64k, 
96k, 128k
& Perplexity & LLaMA-7B & Language modeling, Passkey retrieval & Proof-Pile, GovReport, 
Gutenberg (PG-19), Books3
\label{tab:table_1}

\end{longtable}
\clearpage
\twocolumn

\end{document}